\documentclass{article}
\usepackage{iclr2027_conference,times}

\usepackage{amsmath,amsfonts,bm}

\def\eqref#1{equation~\ref{#1}}

\def\1{\bm{1}}

\DeclareMathAlphabet{\mathsfit}{\encodingdefault}{\sfdefault}{m}{sl}
\SetMathAlphabet{\mathsfit}{bold}{\encodingdefault}{\sfdefault}{bx}{n}

\usepackage{amsmath}
\usepackage{amssymb}
\usepackage{amsthm}

\usepackage{thmtools}

\usepackage{multirow}
\usepackage{booktabs}
\usepackage{colortbl}
\usepackage{wrapfig}
\usepackage{xspace}
\usepackage{caption}
\usepackage{thm-restate}
\usepackage[bottom]{footmisc}
\usepackage{hyperref}
\usepackage{cleveref}
\usepackage{graphicx}
\usepackage{float}

\definecolor{linkblue}{HTML}{1F4E79}
\definecolor{citegreen}{HTML}{4F6F52}
\definecolor{urlbrown}{HTML}{7A4E2D}
\definecolor{notered}{HTML}{A23B3B}
\definecolor{noteblue}{HTML}{2E5C8A}
\definecolor{notemagenta}{HTML}{8A4F7D}

\hypersetup{
  colorlinks=true,
  linkcolor=linkblue,
  citecolor=citegreen,
  urlcolor=urlbrown,
}

\declaretheoremstyle[
  spaceabove=\topsep,
  spacebelow=\topsep,
]{compactthm}
\declaretheorem[style=compactthm,name=Definition]{definition}

\crefname{definition}{Definition}{Definitions}
\Crefname{definition}{Definition}{Definitions}
\crefname{equation}{Equation}{Equations}
\Crefname{equation}{Equation}{Equations}
\crefname{figure}{Figure}{Figures}
\Crefname{figure}{Figure}{Figures}
\crefname{table}{Table}{Tables}
\Crefname{table}{Table}{Tables}
\crefname{section}{Section}{Sections}
\Crefname{section}{Section}{Sections}
\crefname{subsection}{Section}{Sections}
\Crefname{subsection}{Section}{Sections}
\crefname{appendix}{Appendix}{Appendices}
\Crefname{appendix}{Appendix}{Appendices}
\crefname{subappendix}{Appendix}{Appendices}
\Crefname{subappendix}{Appendix}{Appendices}

\AddToHook{cmd/appendix/after}{%
  \crefalias{section}{appendix}%
  \crefalias{subsection}{subappendix}%
}

\newcommand{\OO}{\mathcal{O}}
\newcommand{\Resting}{\mathbf{o}}
\newcommand{\HiddenState}{\mathbf{h}}
\newcommand{\Query}{\mathbf{q}}
\newcommand{\Key}{\mathbf{k}}
\newcommand{\ie}{i.e.\@\xspace}

\title{ReLOBGen: Replayable Limit Order Book \\ Message Generation}
\author{%
Junoh Kang$^{1}$ \qquad\qquad Kiseop Lee$^{2}$ \qquad\qquad Bohyung Han$^{1,3}$ \\
$^{1}$ECE \& $^{3}$IPAI, Seoul National University \qquad $^{2}$Department of Statistics, Purdue University\\
\texttt{junoh.kang@snu.ac.kr} \qquad \texttt{kiseop@purdue.edu} \qquad \texttt{bhhan@snu.ac.kr}
}

\iclrfinalcopy
\begin{document}

\maketitle
\lhead{Preprint}


\begin{abstract}

We propose ReLOBGen, a method for generating limit order book~(LOB) messages that are replayable by construction.
Replayability is required for closed-loop market simulation, yet existing LOB message generators may produce non-replayable raw messages, \ie, messages inconsistent with the current market state.
These generators therefore rely on post-hoc correction or rejection followed by resampling, which may alter the replayed message distribution or increase inference cost.
ReLOBGen instead ensures replayability during generation: it selects the referenced order from the resting orders in the current LOB and then generates the remaining message fields to be consistent with that order and the market state.
For realistic reference selection, ReLOBGen samples from a learned distribution over eligible resting orders, efficiently computed from cached order representations and a context-dependent query.
It then enforces the consistency of the remaining fields by masking out invalid tokens.
Together, these components enable efficient generation of realistic messages without post-hoc correction or resampling.
In 500-message rollouts, ReLOBGen achieves 100\% replayability, improves market realism, particularly for top-of-book statistics and the relative prices of LOB messages, and provides a $2.7\text{--}3.6\times$ speedup per replayed message over the LOBS5 baseline. 
\end{abstract}

\section{Introduction} \label{sec:intro}

\begin{figure}[t]
    \centering
    \includegraphics[width=\textwidth]{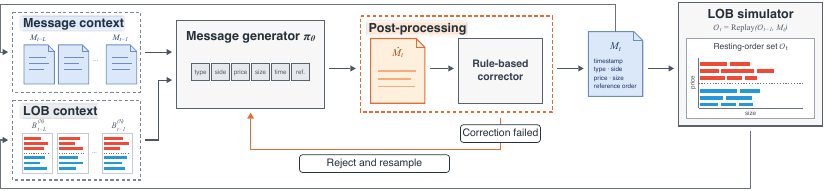}
    \caption{\textbf{Market simulation with an LOB message generator.}
    At each step, the generator produces a raw message $\widehat M_t$.
    When $\widehat M_t$ is not replayable, existing approaches correct it when possible or otherwise reject it and resample.
    ReLOBGen, however, generates replayable messages by construction and requires no post-processing.
    The LOB simulator replays the resulting message $M_t$ to update the market state, which conditions the next generation step.}
    \label{fig:simulation-replayability}
    \vspace{-3mm}
\end{figure}

A limit order book~(LOB) contains buy and sell orders waiting to be traded, called resting orders.
Each resting order is characterized by its price, size, and timestamp.
LOB messages specify changes to the book: add events introduce new resting orders, while cancellation and execution events reduce or remove a referenced resting order.
Recent work~\citep{nagy2023generative,wheeler2024marketgpt} models sequences of LOB messages autoregressively, learning to generate the next message from preceding messages and LOB states.
These models have a range of potential applications in financial markets, such as training and evaluating trading strategies and analyzing market impact~\citep{nagy2023generative,li2025mars}.
To simulate a market for these applications, the model and a rule-based LOB simulator~\citep{byrd2020abides,frey2023jax} operate in a closed loop, as illustrated in \cref{fig:simulation-replayability}.
At each step, the model generates a message, and the simulator updates the LOB according to that message; we refer to this update as \emph{replay}.
The updated LOB then conditions the next generation step.

However, an LOB message generator may produce raw messages that cannot be replayed against the current market state, such as a cancellation message that references a nonexistent resting order or whose size exceeds the referenced order's remaining size.
For example, over 25\% of raw generation attempts in our reproduction of the LOBS5 baseline are not replayable as generated.
Existing approaches therefore rely on method-specific post-processing to obtain a replayable message from a raw output $\widehat M_t$.
This includes correction or rejection followed by resampling if correction fails, as illustrated in the post-processing stage of \cref{fig:simulation-replayability}.
These post-processing procedures are often heuristic: LOBS5~\citep{nagy2023generative} and MarketGPT~\citep{wheeler2024marketgpt} apply different fallback matching rules to identify the order referenced by a cancellation message, whereas MarS~\citep{li2025mars} projects the generated cancellation price to the nearest available resting-order price.
LOBS5 and MarketGPT also interpret the generated cancellation quantity differently: LOBS5 treats it as an absolute amount, whereas MarketGPT preserves the generated cancellation ratio.
Such corrections may alter message semantics and shift the distribution of replayed messages, while resampling increases inference cost.
This problem is structural: replayability depends on the exact current market state, which a finite message context and aggregated LOB states may not fully specify~(see \cref{app:resting_order_coverage}).



We introduce ReLOBGen, which generates replayable LOB messages by construction and thus requires neither post-hoc correction nor resampling.
Its design satisfies two jointly sufficient conditions for replayability: reference eligibility and event-order compatibility---a non-add event must reference an eligible resting order, and the event order must be consistent with the event, the market state, and any reference order it is applied to.
ReLOBGen's message representation encodes the reference order using the target resting order's current attributes, such as price and remaining quantity, so that these attributes can constrain the event-order fields.
During generation, ReLOBGen first generates the event and then selects the reference order from the resting orders eligible for that event, ensuring reference eligibility.
Finally, it generates the event-order fields while masking out invalid token values, ensuring event-order compatibility.
For realistic reference selection, ReLOBGen samples from a learned distribution over eligible resting orders, efficiently computed by dot-product scoring of cached order representations against a context-dependent query.

Empirically, ReLOBGen achieves 100\% replayability, improves rollout realism, and generates messages $2.7\text{--}3.6\times$ faster than the LOBS5 baseline~\citep{nagy2023generative} in 500-message rollouts on GOOG and INTC.
Using the same S5~\citep{smith2022simplified} backbone as the baseline, ReLOBGen achieves this speedup through fewer autoregressive decoding steps and the elimination of resampling.
In terms of rollout realism, ReLOBGen more closely matches the marginal event-type distribution and shows broad improvements across unconditional, conditional, and market-impact evaluations with LOB-Bench~\citep{nagy2025lob}.
These improvements are particularly pronounced in metrics capturing top-of-book statistics and the prices of submitted and canceled orders relative to the current book.
An ablation shows that learned reference selection improves rollout realism over uniform selection from the same eligible resting orders.

Overall, ReLOBGen makes every generated message replayable by construction---selecting reference orders from the eligible resting orders and using the selected orders to constrain event-order generation.
Our results show that this guarantee can be achieved alongside improved rollout realism and computational efficiency.
Generated messages are replayed without modification, preserving their semantics and avoiding distribution changes introduced by method-specific correction rules.
Together, these results establish ReLOBGen as a practical method for efficiently generating replayable and realistic LOB messages.

\section{Related work}
\label{sec:rel_work}

\subsection{Modeling financial market microstructure}
\label{subsec:rel_work_modeling}

Data-driven modeling of financial markets spans multiple targets, from future price movements to the evolution of the limit order book.
DeepLOB~\citep{zhang2019deeplob} predicts future mid-price movements from recent LOB observations.
Because the mid-price provides only a compressed representation of market state, subsequent work directly forecasts or generates the joint evolution of prices and queued volumes across multiple book levels~\citep{jung2025attention,backhouse2025painting}.

Another line of work models order flow---the sequence of events that update the book---using stochastic processes.
\citet{smith2003statistical} treat order flow as independent random events and analyze continuous double-auction statistics, while \citet{cont2010stochastic} model limit, market, and cancellation events as Poisson flows calibrated to high-frequency data.
\citet{huang2015simulating} make event intensities depend on the current book state in a queue-reactive model, whereas \citet{bacry2015hawkes} describe multivariate Hawkes processes that capture temporal clustering and self- and cross-excitation through history-dependent intensities.

More recent work uses neural autoregressive models to generate order flow.
LOBS5~\citep{nagy2023generative} uses an S5-based state-space model~\citep{smith2022simplified} to generate tokenized LOB messages from LOBSTER data~\citep{Huang2011LOBSTER} for two Nasdaq equities, conditioned on past messages and LOB states.
MarketGPT~\citep{wheeler2024marketgpt} uses a decoder-only Transformer~\citep{vaswani2017attention} to generate tokenized ITCH messages from message histories, with full-depth Nasdaq pretraining across 20 equities and asset-specific fine-tuning.
MarS~\citep{li2025mars} combines order-level and order-batch models to generate orders conditioned on past orders and LOB states, training on 500 liquid Chinese equities.
TradeFM~\citep{kawawa2026tradefm} scales Transformer-based order-flow generation to over 9,000 US equities, conditioning on LOB message streams.


\subsection{Evaluating message-generation models}
\label{subsec:rel_work_evaluating}

Message-generation models are commonly evaluated using next-token perplexity and market realism in autoregressive rollouts~\citep{nagy2023generative,kawawa2026tradefm}.
LOB-Bench~\citep{nagy2025lob} assesses realism by comparing distributions of market microstructure statistics, such as spreads and message inter-arrival times, extracted from real and generated messages and LOB trajectories.
The 21 metrics span six groups: top-of-book conditions~(\emph{State}), event timing~(\emph{Times}), resting liquidity~(\emph{Volumes}), the locations of order submissions and cancellations, measured by price distance and book-level rank~(\emph{Depths} and \emph{Levels}, respectively), and trading activity and order-flow imbalance~(\emph{Trades}).
These comparisons cover unconditional and conditional distributions, as well as error accumulation measured by changes in distributional divergence over the rollout horizon.
It also evaluates market impact through price responses to order events and temporal correlations between order events.
Additionally, we evaluate raw-message replayability by measuring how often generated messages require correction or rejection and whether they violate reference eligibility or event-order compatibility.

\section{Problem formulation}\label{sec:problem_formulation}

\subsection{LOB message generation model}\label{subsec:message_generation_simulation}

We first define the main objects used in LOB message generation.
Let $\mathcal O_{t-1}$ denote the set of resting orders at step $t-1$.
An LOB message $M_t=(E_t,R_t,X_t)$ describes a change to this set: $E_t$ specifies the event type and side~(bid or ask), $R_t$ denotes the reference order, and $X_t$ denotes the event order.
We consider four event types: add inserts a new resting order, cancel partially reduces a referenced order's remaining quantity, delete removes the referenced order entirely, and execute reduces a referenced order's remaining quantity through a trade.
The orders involved in these events are represented by $X_t$ and $R_t$, each containing price, size, and timestamp fields.
For add events, $X_t$ describes the newly submitted order, and no reference order is required.
For non-add events, $R_t$ identifies the referenced resting order, and $X_t$ specifies the price and quantity of the operation on that order.
Replaying $M_t$ updates the resting-order set to $\mathcal O_t$.
Aggregating the remaining quantities in $\mathcal O_t$ by side and price and retaining the best $N$ occupied price levels on each side yields the $N$-level LOB state $B_t^{(N)}$, which does not retain individual resting orders.

An LOB message generation model $\pi_\theta$ learns the conditional distribution of the next message $M_t$ given the preceding $L$ LOB messages $M_{t-L:t-1}$ and $N$-level LOB states $B_{t-L:t-1}^{(N)}$.
Each message is tokenized and modeled autoregressively using architectures such as Transformers~\citep{vaswani2017attention} or state-space models~\citep{smith2022simplified}.
The generation model used in this work follows LOBS5~\citep{nagy2023generative}, using an S5 architecture and representing each message $M_t$ as a sequence of 22 field-wise tokens $z_{t,1:22}$~(see \cref{tab:normalization_encoding}).
The model is trained to minimize the standard autoregressive negative log-likelihood:
\begin{align}
    \mathcal L_{\mathrm{NLL}}(\theta)
    = -\mathbb E_{t,k} \left[
        \log \pi_\theta\left(
            z_{t,k} \mid M_{t-L:t-1},B_{t-L:t-1}^{(N)},z_{t,<k}
        \right) 
    \right].
\end{align}





\begin{table}[t]
  \centering
  \caption{\textbf{Event-specific replayability conditions.}
  Here, $\mathrm{side}=E_t[\mathrm{side}]$, $\mathcal O_{t-1}^{(\mathrm{side})}$ contains the resting orders on that side, and $R_t$ carries the referenced order's current attributes.
  $\operatorname{BestAsk}$ and $\operatorname{BestBid}$ denote the lowest ask and highest bid prices in the current LOB state.
  $\operatorname{FrontOfQueue}$ returns the oldest order resting at the best price on that side.
  Under the LOB simulator's price--time priority, an execute event allows only this order as its reference.}
  \label{tab:field_validity_conditions}

  \renewcommand{\arraystretch}{1.4}
  \setlength{\tabcolsep}{4pt}
  \scalebox{0.90}{%
    \begin{tabular}{ccl}
      \toprule
      Event type & Eligible resting-order set $\mathcal C_t$ & Event-order constraints \\
      \midrule
      Add &
      -- &
      Bid: $X_t[\mathrm{price}] < \operatorname{BestAsk}$;
      Ask: $X_t[\mathrm{price}] > \operatorname{BestBid}$. \\
      Cancel &
      $\{o\in\mathcal O_{t-1}^{(\mathrm{side})}:o[\mathrm{size}]>1\}$ &
      $X_t[\mathrm{price}]=R_t[\mathrm{price}]$ and
      $1\le X_t[\mathrm{size}]<R_t[\mathrm{size}]$. \\
      Delete &
      $\{o\in\mathcal O_{t-1}^{(\mathrm{side})}:o[\mathrm{size}]>0\}$ &
      $X_t[\mathrm{price}]=R_t[\mathrm{price}]$ and
      $X_t[\mathrm{size}]=R_t[\mathrm{size}]$. \\
      Execute &
      $\{\operatorname{FrontOfQueue}(\mathcal O_{t-1}^{(\mathrm{side})})\}$ &
      $X_t[\mathrm{price}]=R_t[\mathrm{price}]$ and
      $1\le X_t[\mathrm{size}]\le R_t[\mathrm{size}]$. \\
      \bottomrule
    \end{tabular}%
  }
\end{table}


\subsection{Market simulation and replayability}\label{subsec:simulation_and_replayability}

To simulate a market trajectory autoregressively, the message generation model $\pi_\theta$ and the LOB simulator alternate between message generation and replay, as illustrated in \cref{fig:simulation-replayability}.
At each step, the model generates a message conditioned on the preceding message and LOB state histories.
The simulator replays the message $M_t$ against the current resting-order set:
\begin{align}
    \mathcal O_t &= \operatorname{Replay}(\mathcal O_{t-1},M_t),
\end{align}
where $\operatorname{Replay}$ denotes the event-specific deterministic state-transition operator used by the LOB simulator.
The message $M_t$ and the LOB state $B_t^{(N)}$ aggregated from $\mathcal O_t$ are appended to their respective histories to condition the next generation step.

However, a message $\widehat M_t$ generated by $\pi_\theta$ may not be replayable against the current resting-order set $\mathcal O_{t-1}$.
For example, a non-add event may reference an order that is absent from $\mathcal O_{t-1}$ or ineligible for that event.
Even with an eligible reference, the event order may violate event-specific constraints, such as exceeding the reference order's remaining quantity; an add event may instead violate the price constraint imposed by the current market state.
In our reproduction of LOBS5~(ref.-last), over 25\% of raw generation attempts require correction or rejection on both stocks~(\cref{tab:full_model_replayability}).
The larger MarketGPT model also reports that approximately 7\% of generated messages cannot be corrected and require resampling~\citep{wheeler2024marketgpt}.
Based on the LOB simulator's replay rules, we define replayability as follows.

\begin{definition}[Replayability]\label{def:replayability}
A message $M_t=(E_t,R_t,X_t)$ is replayable with respect to the current resting-order set $\mathcal O_{t-1}$ if it satisfies both conditions:
\begin{itemize}
    \item \textbf{Reference eligibility.} For a non-add event, the reference order $R_t$ must belong to the eligible resting-order set $\mathcal C_t(E_t,\mathcal O_{t-1})$ defined in \cref{tab:field_validity_conditions}.
    \item \textbf{Event-order compatibility.} The event order $X_t$ satisfies the constraints in \cref{tab:field_validity_conditions}.
\end{itemize}
\end{definition}

When a generated message $\widehat M_t$ is not replayable, existing approaches use method-specific post-processing to obtain a replayable message $M_t$~\citep{nagy2023generative,wheeler2024marketgpt,li2025mars}.
This includes correction or rejection followed by resampling, with event-specific procedures detailed in~\cref{app:post-process}.
Although this post-processing makes $M_t$ replayable, it introduces two limitations.
First, method-specific corrections may modify generated fields or reinterpret their semantics, shifting the distribution of replayed messages away from that of raw model outputs.
Second, resampling rejected messages increases inference cost.
These limitations motivate generating messages that are replayable by construction while preserving rollout realism.

\section{ReLOBGen: Replayable LOB message generation}
\label{sec:method}

ReLOBGen generates each LOB message by first generating the event $E_t$, then resolving the reference order $R_t$, and finally generating the event order $X_t$.
For non-add events, resolving the referenced resting order first allows the event order's price and quantity to be generated within the constraints imposed by that order, which makes reference-first generation a natural choice.
We adapt the message representation to support this reference-first generation process~(\cref{subsec:reference_first_factorization}).
During generation, ReLOBGen first generates the event $E_t$.
For non-add events, ReLOBGen then selects the reference order from the eligible resting-order set~(\cref{subsec:reference_order_existence}).
Finally, it generates the event order to be consistent with the event, any selected reference order, and the current market state by masking out invalid token values~(\cref{subsec:field_validity}).
Together, these mechanisms produce replayable messages by construction without post-hoc correction or resampling.

\begin{figure}[t]
    \centering
    \begin{tabular}{@{}c@{\hspace{6pt}{\color[HTML]{D1D5DB}\vrule width\arrayrulewidth}\hspace{6pt}}c@{}}
        \multicolumn{2}{@{}c@{}}{\includegraphics[width=\textwidth]{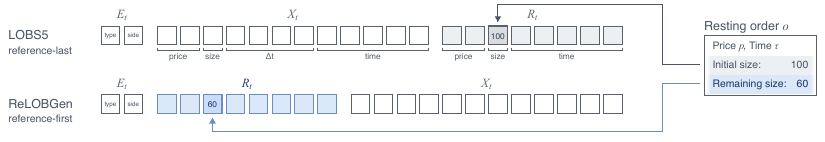}} \\
        \multicolumn{2}{@{}c@{}}{(a) Reference-first message representation} \\[8pt]
        \includegraphics[width=0.58\textwidth]{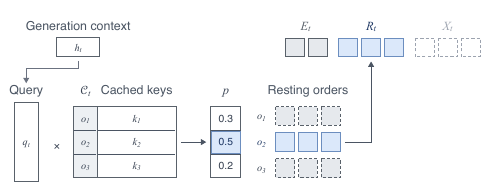} &
        \includegraphics[width=0.385\textwidth]{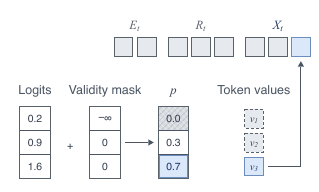} \\
        (b) Resting-order selection & (c) Inference-time masking
    \end{tabular}
    \caption{\textbf{ReLOBGen: mechanisms for replayability by construction.}
    (a) The message representation places $R_t$ before $X_t$ and encodes the referenced order's remaining size rather than its initial submitted size.
    (b) ReLOBGen samples $R_t$ from a learned distribution over the eligible resting-order set $\mathcal C_t$.
    This distribution is efficiently computed by applying a softmax to the scaled dot products between a query derived from the generation context~(the hidden state after processing $E_t$) and the cached order keys.
    (c) ReLOBGen masks invalid token values and samples from the remaining valid support during $X_t$ generation.}
    \label{fig:relobgen}
\end{figure}

\subsection{Reference-first message representation}\label{subsec:reference_first_factorization}

For ReLOBGen, we make two modifications to the LOBS5~\citep{nagy2023generative} message representation, as illustrated in \cref{fig:relobgen}~(a), while retaining its field-wise tokenization and vocabulary~(see \cref{tab:normalization_encoding} in \cref{app:data_preparation}).
First, we place $R_t$ before $X_t$ in the token sequence, yielding the order $E_t \rightarrow R_t \rightarrow X_t$ and making the reference available to determine the constraints on $X_t$ before its generation.
Second, we represent $R_t$ using the referenced resting order's current attributes so that these constraints reflect the current market state.
Specifically, we set $R_t[\mathrm{size}]$ to the order's remaining size in $\OO_{t-1}$ rather than the initial submitted size used by LOBS5.
We train the message generation model $\pi_\theta$ on this representation using the negative log-likelihood objective in \cref{subsec:message_generation_simulation}.

\subsection{Resting-order selection for reference eligibility}\label{subsec:reference_order_existence}

\paragraph{Resting-order selection.}
ReLOBGen guarantees the reference eligibility condition in \cref{def:replayability} by selecting $R_t$ only from the eligible resting-order set $\mathcal C_t(E_t,\OO_{t-1})$ specified in \cref{tab:field_validity_conditions}.
Note that the number of model forward passes per message is reduced by selecting $R_t$ rather than autoregressively generating its eight tokens, and non-add events with no eligible resting orders are masked during event generation.

While selecting any eligible resting order guarantees reference eligibility, the particular choice still affects the realism of the generated LOB messages.
ReLOBGen therefore samples $R_t$ from a learned categorical distribution $p_\phi(\cdot\mid\HiddenState_t,\mathcal C_t)$ over the eligible resting orders, where $\HiddenState_t$ denotes the generation context up to and including $E_t$.
With the base model frozen, we train this distribution by minimizing the cross-entropy for the observed reference order $R_t^\star$:
\begin{align}
    \mathcal L_{\mathrm{sel}}
    = -\mathbb E_t\left[\log p_\phi(R_t^\star\mid\HiddenState_t,\mathcal C_t)\right].
\end{align}

\paragraph{Efficient resting-order selection with cached keys.}
ReLOBGen implements resting-order selection efficiently by exploiting a structural property of LOB messages: each message affects only one resting-order entry.
We therefore cache a key $\Key_o$ for each resting order and reuse the keys of unchanged orders.
Using query--key scoring similar to dense retrieval~\citep{karpukhin2020dense}, each eligible resting order $\Resting\in\mathcal C_t(E_t,\OO_{t-1})$ is scored by the scaled dot product between its cached key $\Key_o$ and a query $\Query_t$ representing the current generation context, as illustrated in \cref{fig:relobgen}~(b):
\begin{align}
    s_\phi(\Resting,\HiddenState_t) = \Query_t^\top \Key_o / \sqrt d,
\end{align}
where $d$ is the key and query dimension.
Applying a softmax over these scores within the eligible resting-order set yields $p_\phi(\cdot\mid\HiddenState_t,\mathcal C_t)$.

To keep cached keys reusable as the market moves, we encode resting orders relative to a fixed mid-price anchor rather than the current mid-price, while the query incorporates the current generation context and the displacement of the current mid-price from this anchor.
We represent generation context $\HiddenState_t$ using the base model's hidden state and compute the order key and shared query using lightweight MLPs, $f_{\mathrm{order}}$ and $f_{\mathrm{query}}$:
\begin{align}
    \Key_o &= f_{\mathrm{order}}(\Resting; \mathrm{mid}_{\mathrm{anchor}}),
    &\Query_t &= f_{\mathrm{query}}(\HiddenState_t, \mathrm{mid}_{t-1}; \mathrm{mid}_{\mathrm{anchor}}).
\end{align}
Here, $\mathrm{mid}_{\mathrm{anchor}}$ is the fixed mid-price anchor, and $\mathrm{mid}_{t-1}$ is the mid-price at step $t-1$.

At the start of each rollout, we set $\mathrm{mid}_{\mathrm{anchor}}$ to the initial mid-price.
When an order is added or updated, we compute its key from the order's current attributes; when it is removed, we discard its key; we maintain cached keys for unaffected orders.
When training $f_{\mathrm{order}}$ and $f_{\mathrm{query}}$, we sample $\mathrm{mid}_{\mathrm{anchor}}$ from earlier mid-prices in the context to expose them to price drift relative to the anchor.

\subsection{Inference-time masking for event-order compatibility}\label{subsec:field_validity}

ReLOBGen guarantees event-order compatibility by restricting each field in $X_t$ to the valid support determined by $E_t$, $R_t$, and the current market state, as specified in \cref{tab:field_validity_conditions}.
\cref{fig:relobgen}~(c) illustrates this process: the model masks invalid tokens and samples from the remaining support, as in constrained decoding for structured generation~\citep{scholak2021picard}.
When the valid support is a singleton, ReLOBGen directly assigns the unique valid token, further reducing the number of autoregressive model forward passes.
The pretrained model assigns negligible probability mass to invalid tokens on teacher-forced test windows---less than $0.14\%$ on GOOG and $0.29\%$ on INTC, suggesting a limited impact of forcing and masking.


\section{Experiments}
\label{sec:experiments}

\subsection{Experimental setup}
\label{subsec:experimental_setup}

\paragraph{Data and model training.}
We use market-by-order~(MBO) data for GOOG and INTC from Databento\footnote{\label{fn:databento}\url{https://databento.com/portal/catalog/us-equities\#XNAS.ITCH}}, with 2025 data for training and separate periods in January 2026 for validation and testing.
We follow the preprocessing and field-wise tokenization of LOBS5~\citep{nagy2023generative}, except for the reference-field semantics described in~\cref{subsec:reference_first_factorization}~(see \cref{app:data_preparation} for details).
We train 35M-parameter models using the same S5 backbone as the scaled-up LOBS5 in \citet{nagy2025lob}, with reference-last and reference-first token orders.
We additionally train two lightweight MLP projection heads with 2.3M parameters in total for learned reference selection~(see \cref{app:model_training} for training details).

\paragraph{Rollout evaluation.}
We compare the following configurations on the same 1,000 rollout windows per stock, with 500 successfully replayed messages per rollout, matching the rollout length reported in LOB-Bench~\citep{nagy2025lob}~(see \cref{app:rollout_evaluation} for details).
ReLOBGen combines the reference-first model with learned selection from the eligible resting-order set and inference-time masking.
Our baselines autoregressively generate messages with either model and apply the LOBS5 post-processing procedure~(see \cref{app:lobs5} for details); we denote them by LOBS5~(ref.-last) and LOBS5~(ref.-first).
Ref.-first + uniform uses the reference-first model and uniform reference selection with inference-time masking.
All configurations start from the same exact resting-order set, which is maintained by the JAX-LOB simulator~\citep{frey2023jax} and available to both ReLOBGen and baseline post-processing.
We assess replayability on raw generation attempts before post-processing and evaluate rollout realism from replayed messages and resulting LOB trajectories using LOB-Bench and the marginal event-type distribution.



\begin{table}[t]
  \centering
  \caption{\textbf{Post-processing rates, replayability violations, and runtime.}
  Runtime is measured on an NVIDIA RTX A6000 GPU and reported as mean $\pm$ standard deviation.
  Aborted rollouts---59 and 17 for LOBS5~(ref.-last) and LOBS5~(ref.-first) on GOOG, respectively, and two for each baseline on INTC---are excluded from evaluation.
  Neither ReLOBGen nor Ref.-first + uniform requires any restarts on either stock.
  Lower is better, and bold indicates the lowest value.}
  \label{tab:full_model_replayability}
  \renewcommand{\arraystretch}{1.1}
  \setlength{\tabcolsep}{3pt}
  \scalebox{0.85}{
    \begin{tabular}{@{}clcc>{\hspace{8pt}}cc>{\hspace{8pt}}cc@{}}
      \toprule
      & & \multicolumn{2}{c}{Post-processing} & \multicolumn{2}{c}{Replayability violations} & \multicolumn{2}{c}{Runtime (ms)} \\
      \cmidrule(lr){3-4}\cmidrule(lr){5-6}\cmidrule(lr){7-8}
      Ticker & Method & Correction & Rejection & Reference & Event-order & Per replayed msg. & Per attempt \\
      \midrule
      \multirow{3}{*}{GOOG}
      & LOBS5~(ref.-last) & 13.5\% & 25.7\% & 39.1\% & 0.4\% & $602 \pm 164$ & $447 \pm 119$ \\
      & LOBS5~(ref.-first) & 11.7\% & 27.6\% & 39.5\% & 0.1\% & $591 \pm 145$ & $426 \pm 100$ \\
      & \cellcolor{gray!10}ReLOBGen & \cellcolor{gray!10}\textbf{0.0\%} & \cellcolor{gray!10}\textbf{0.0\%} & \cellcolor{gray!10}\textbf{0.0\%} & \cellcolor{gray!10}\textbf{0.0\%} & \cellcolor{gray!10}$\mathbf{227 \pm 12}$ & \cellcolor{gray!10}$\mathbf{227 \pm 12}$ \\
      \midrule
      \multirow{3}{*}{INTC}
      & LOBS5~(ref.-last) & 9.7\% & 16.3\% & 25.6\% & 0.5\% & $1216 \pm 726$ & $1018 \pm 606$ \\
      & LOBS5~(ref.-first) & 11.5\% & 13.2\% & 24.5\% & 0.2\% & $1058 \pm 850$ & $921 \pm 744$ \\
      & \cellcolor{gray!10}ReLOBGen & \cellcolor{gray!10}\textbf{0.0\%} & \cellcolor{gray!10}\textbf{0.0\%} & \cellcolor{gray!10}\textbf{0.0\%} & \cellcolor{gray!10}\textbf{0.0\%} & \cellcolor{gray!10}$\mathbf{341 \pm 150}$ & \cellcolor{gray!10}$\mathbf{341 \pm 150}$ \\
      \bottomrule
    \end{tabular}
  }
\end{table}

\subsection{Replayability}
\label{subsec:replayability}
\cref{tab:full_model_replayability} summarizes post-processing and replayability violation rates, with an event-wise breakdown in \cref{app:eventwise_replayability}.
For non-add events, event-order compatibility is assessed against the generated reference fields, regardless of whether the reference is eligible.
Both LOBS5 baselines frequently require correction or rejection, with reference eligibility violations far more frequent than event-order compatibility violations.
This gap may partly reflect the difficulty of identifying eligible resting orders, many of which are not individually specified by the finite message context or the aggregated LOB state, as analyzed in \cref{app:resting_order_coverage}.
Both baselines also require rollout restarts after 100 consecutive failed message-generation attempts.
On the other hand, ReLOBGen satisfies both replayability conditions by construction, recording zero violations and requiring no post-processing throughout the evaluated rollouts.
Note that LOBS5~(ref.-last) still requires post-processing when its resting-order set is initialized with INIT orders, as shown in \cref{app:init_fallback}.

Relative to LOBS5~(ref.-last), ReLOBGen achieves $2.7\times$ and $3.6\times$ speedups in runtime per replayed message on GOOG and INTC, respectively.
Runtime covers the full closed-loop simulation, including message generation, post-processing when required, and simulator state updates.
These gains come from reduced autoregressive decoding and the elimination of post-processing.
Selecting the reference order rather than generating its tokens autoregressively and forcing singleton-valid field tokens reduces non-add decoding from 17 to 7--8 model forward passes.
Eliminating rejection also removes the $1/(1-r)$ factor in attempts per replayed message, where $r$ denotes the rejection rate.
The larger speedup on INTC might reflect higher reference-matching costs in the baseline due to its larger resting-order population.



\begin{table}[t]
  \centering
  \caption{\textbf{LOB-Bench evaluation: metric-group summary of unconditional $L_1$ distances.}
  Overall averages all metrics, while each metric-group column averages the metrics in its group. 
  Lower is better, and bold indicates the lowest value.}
  \label{tab:lob_bench_unconditional_summary_l1}
  \renewcommand{\arraystretch}{1.1}
  \setlength{\tabcolsep}{3.5pt}
  \scalebox{0.90}{%
    \begin{tabular}{@{}cl@{\hspace{5pt}}ccccccc@{}}
      \toprule
      & & & \multicolumn{6}{c}{Metric group} \\
      \cmidrule(lr){4-9}
      Ticker & Method & Overall & State & Times & Volumes & Depths & Levels & Trades \\
      \midrule
      \multirow{3}{*}{GOOG}
      & LOBS5~(ref.-last) & 0.27 & 0.35 & 0.16 & 0.25 & 0.40 & 0.21 & 0.23 \\
      & Ref.-first + uniform & 0.24 & 0.31 & 0.20 & 0.22 & 0.23 & 0.20 & 0.27 \\
      & \cellcolor{gray!10}ReLOBGen & \cellcolor{gray!10}\textbf{0.17} & \cellcolor{gray!10}\textbf{0.24} & \cellcolor{gray!10}\textbf{0.15} & \cellcolor{gray!10}\textbf{0.18} & \cellcolor{gray!10}\textbf{0.20} & \cellcolor{gray!10}\textbf{0.11} & \cellcolor{gray!10}\textbf{0.18} \\
      \midrule
      \multirow{3}{*}{INTC}
      & LOBS5~(ref.-last) & 0.22 & 0.14 & \textbf{0.14} & 0.20 & 0.32 & 0.17 & 0.27 \\
      & Ref.-first + uniform & 0.30 & 0.10 & 0.35 & 0.27 & 0.36 & 0.34 & 0.32 \\
      & \cellcolor{gray!10}ReLOBGen & \cellcolor{gray!10}\textbf{0.11} & \cellcolor{gray!10}\textbf{0.05} & \cellcolor{gray!10}0.21 & \cellcolor{gray!10}\textbf{0.16} & \cellcolor{gray!10}\textbf{0.07} & \cellcolor{gray!10}\textbf{0.04} & \cellcolor{gray!10}\textbf{0.15} \\
      \bottomrule
    \end{tabular}
  }
\end{table}


\begin{table}[t]
  \centering
  \caption{\textbf{LOB-Bench evaluation: conditional $L_1$ distances and market-impact discrepancies.}
  Market-impact discrepancies cover market orders~(MO), limit orders~(LO), and cancellations~(CA), with subscripts 0 and 1 for events without and with an immediate mid-price change.
  Conditional 99\% percentile bootstrap confidence intervals have half-widths below 0.004.
  Lower is better, and bold indicates the lowest value.}
  \label{tab:lob_bench_cond_impact}
  \renewcommand{\arraystretch}{1.1}
  \setlength{\tabcolsep}{3pt}
  \scalebox{0.80}{
    \begin{tabular}{@{}llccc>{\hspace{6pt}}cccccc}
      \toprule
      & & \multicolumn{3}{c}{Conditional ($L_1$)} & \multicolumn{6}{c}{Market impact} \\
      \cmidrule(lr){3-5}\cmidrule(lr){6-11}
      Ticker & Method
      & ask vol. $\mid$ spread & spread $\mid$ time & spread $\mid$ volatility
      & $MO_0$ & $MO_1$ & $LO_0$ & $LO_1$ & $CA_0$ & $CA_1$ \\
      \midrule
      \multirow{3}{*}{GOOG} & LOBS5~(ref.-last)
      & \textbf{0.14} & 0.63 & 0.46
      & 23.4 & 16.9 & 5.5 & 8.7 & 4.5 & \textbf{4.7} \\
      & Ref.-first + uniform
      & 0.16 & 0.51 & 0.57
      & 12.1 & 13.9 & 10.5 & 4.9 & 7.6 & 8.8 \\
      & \cellcolor{gray!10}ReLOBGen
      & \cellcolor{gray!10}0.18 & \cellcolor{gray!10}\textbf{0.37} & \cellcolor{gray!10}\textbf{0.42}
      & \cellcolor{gray!10}\textbf{7.2} & \cellcolor{gray!10}\textbf{10.4} & \cellcolor{gray!10}\textbf{1.7} & \cellcolor{gray!10}\textbf{3.1} & \cellcolor{gray!10}\textbf{3.3} & \cellcolor{gray!10}5.3 \\
      \midrule
      \multirow{3}{*}{INTC} & LOBS5~(ref.-last)
      & 0.22 & 0.22 & 0.16
      & 9.9 & 5.9 & 2.3 & 3.5 & \textbf{0.4} & 11.8 \\
      & Ref.-first + uniform
      & 0.49 & 0.15 & 0.09
      & 7.2 & \textbf{5.4} & \textbf{0.8} & \textbf{2.2} & 1.3 & 11.6 \\
      & \cellcolor{gray!10}ReLOBGen
      & \cellcolor{gray!10}\textbf{0.20} & \cellcolor{gray!10}\textbf{0.05} & \cellcolor{gray!10}\textbf{0.06}
      & \cellcolor{gray!10}\textbf{6.2} & \cellcolor{gray!10}6.3 & \cellcolor{gray!10}\textbf{0.8} & \cellcolor{gray!10}\textbf{2.2} & \cellcolor{gray!10}0.5 & \cellcolor{gray!10}\textbf{11.2} \\
      \bottomrule
    \end{tabular}
  }
\end{table}

\subsection{Rollout realism}
\label{subsec:rollout_realism}

Beyond generating fully replayable messages, ReLOBGen also improves rollout realism over the LOBS5 baseline, with broadly lower unconditional $L_1$ distances across metric groups, as shown in \cref{tab:lob_bench_unconditional_summary_l1}.
These improvements are particularly pronounced in the State, Depths, and Levels metric groups.
State captures top-of-book statistics, while Depths and Levels characterize the locations of order submissions and cancellations, measured by price distance and book-level rank, respectively.
At the individual-metric level, ReLOBGen achieves lower point estimates than LOBS5 on 18 of 21 metrics for GOOG, with statistically significant improvements on 16 of those 18.
On INTC, it achieves lower point estimates on 20 of 21 metrics, with statistically significant improvements on 18 of those 20.
Statistical significance is assessed using non-overlapping 99\% bootstrap confidence intervals.
Detailed $L_1$ and Wasserstein results are shown in \cref{fig:lob-bench-uncond,fig:lob_bench_unconditional_w1}, respectively, with metric-group averages for Wasserstein distances reported in \cref{tab:lob_bench_unconditional_summary_w1}, all in \cref{app:additional_lob_bench}.

\begin{figure}[t]
  \centering
  \setlength{\tabcolsep}{1pt}
  \renewcommand{\arraystretch}{0.0}
  \begin{tabular}{@{}cccc@{}}
    \multicolumn{4}{c}{\includegraphics[width=0.35\textwidth]{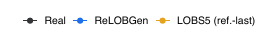}} \\
    \includegraphics[width=0.24\textwidth]{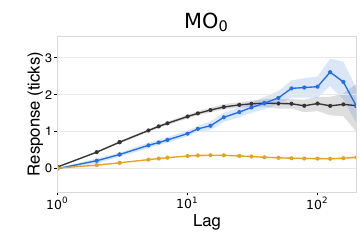} &
    \includegraphics[width=0.24\textwidth]{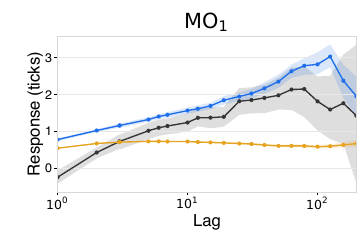} &
    \includegraphics[width=0.24\textwidth]{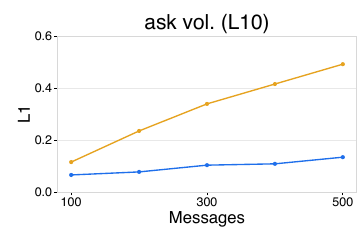} &
    \includegraphics[width=0.24\textwidth]{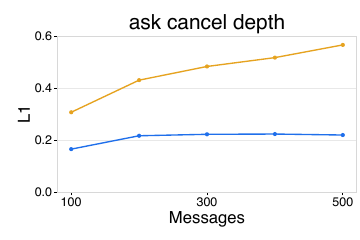} \\
    \multicolumn{2}{c}{\sffamily\fontsize{6}{7}\selectfont (a) Market impact} &
    \multicolumn{2}{c}{\sffamily\fontsize{6}{7}\selectfont (b) Error accumulation} \\
  \end{tabular}
  \caption{\textbf{LOB-Bench evaluation: market-impact response curves and error accumulation on GOOG.}
  (a) Responses after market orders without~($MO_0$) and with~($MO_1$) an immediate mid-price change; shading shows 99\% bootstrap confidence intervals.
  (b) $L_1$ distance over successive 100-message intervals.}
  \label{fig:lob_bench_selected_dynamics}
\end{figure}

\cref{tab:lob_bench_cond_impact} shows that ReLOBGen achieves lower average conditional and market-impact discrepancies than LOBS5 on both stocks, reducing conditional averages~(0.41 to 0.32 on GOOG; 0.20 to 0.10 on INTC) and market-impact averages~(10.6 to 5.2 on GOOG; 5.6 to 4.5 on INTC).
\cref{fig:lob_bench_selected_dynamics}~(a) further illustrates the difference in lagged price-response curves: LOBS5 tends to produce largely flat responses, whereas ReLOBGen better captures those observed in the real data.
Regarding error accumulation, \cref{fig:lob_bench_selected_dynamics}~(b) shows lower errors for ReLOBGen on the two illustrated metrics, suggesting potential for more stable generation over longer rollout horizons.
ReLOBGen also better preserves the marginal event-type distribution, reducing the total variation distance from 9.6 to 6.2 percentage points on GOOG, as detailed in~\cref{tab:marginal_event_distribution} in \cref{app:marginal_event_distribution}.
Additional LOB-Bench evaluation results, including conditional Wasserstein results~(\cref{tab:lob_bench_cond_impact_w1}), the full market-impact response curves~(\cref{fig:lob_bench_impact_response}), and error accumulation for all metrics~(\cref{fig:lob_bench_divergence_l1,fig:lob_bench_divergence_w1}), are provided in \cref{app:additional_lob_bench}.


\begin{table}[t]
  \centering
  \caption{\textbf{Teacher-forced NLLs under reference-last and reference-first token orders.}
  For fields represented by multiple tokens, NLLs are summed across tokens.
  Shaded columns highlight the price and size fields of the reference and event orders.
  Lower is better.}
  \label{tab:teacher_forced_nll_full}
  \renewcommand{\arraystretch}{1.1}
  \setlength{\tabcolsep}{3pt}
  \scalebox{0.85}{
    \begin{tabular}{ccccc>{\columncolor{gray!10}}c>{\columncolor{gray!10}}ccc>{\columncolor{gray!10}}c>{\columncolor{gray!10}}ccc}
      \toprule
      & & & \multicolumn{2}{c}{$E_t$} & \multicolumn{4}{c}{$R_t$} & \multicolumn{4}{c}{$X_t$} \\
      \cmidrule(lr){4-5}\cmidrule(lr){6-9}\cmidrule(lr){10-13}
      Ticker & Token order & Overall & Type & Side & Price & Size & Time (s) & Time (ns) & Price & Size & $\Delta t$ (s) & $\Delta t$ (ns) \\
      \midrule
      \multirow{2}{*}{GOOG} & Ref.-last & 19.01 & 0.76 & 0.51 & 0.00 & 0.02 & 0.29 & 1.33 & 1.83 & 1.29 & 0.00 & 12.96 \\
      & Ref.-first & 18.94 & 0.75 & 0.51 & 0.89 & 0.55 & 0.30 & 1.29 & 0.96 & 0.74 & 0.00 & 12.97 \\
      \midrule
      \multirow{2}{*}{INTC} & Ref.-last & 18.50 & 0.71 & 0.51 & 0.00 & 0.06 & 0.38 & 1.65 & 0.97 & 1.95 & 0.01 & 12.27 \\
      & Ref.-first & 18.51 & 0.72 & 0.50 & 0.46 & 0.82 & 0.37 & 1.66 & 0.51 & 1.20 & 0.01 & 12.28 \\
      \bottomrule
    \end{tabular}
  }
  \vspace{-3mm}
\end{table}


\subsection{Ablation studies}
\label{subsec:ablation}

\paragraph{Effect of learned reference selection.}
Although uniform selection generates replayable messages, it does not necessarily improve rollout realism.
As shown in \cref{tab:lob_bench_unconditional_summary_l1}, it only slightly reduces the overall $L_1$ distance relative to LOBS5~(ref.-last) on GOOG and even increases it on INTC, from 0.22 to 0.30.
Thus, which eligible order is selected matters for rollout realism.

\cref{tab:lob_bench_unconditional_summary_l1,tab:lob_bench_cond_impact} show that learned selection improves overall rollout realism over uniform selection on both stocks.
Learned selection improves reference-order choices for cancel and delete events, with lower $L_1$ distances on all five related metrics for both stocks, as shown in \cref{fig:lob-bench-uncond} in \cref{app:additional_lob_bench}: log time-to-cancel and the bid- and ask-side cancellation depth and level distributions.
The gains are especially pronounced on INTC, where the top ten price levels of the LOB contain 2.7 times as many resting orders as in GOOG, consistent with the potential benefit of learned over uniform selection in larger eligible sets.

\paragraph{Effect of reference-first factorization.}

We evaluate field-wise teacher-forced NLLs under the two token orders on 10,000 windows from the test split.
\cref{tab:teacher_forced_nll_full} shows similar overall NLLs and similar field-wise NLLs for fields other than price and size.
The price and size NLLs suggest that reference-first ordering redistributes predictive uncertainty from the event-order fields to the reference-order fields that now precede them: reference-order NLLs increase, while the corresponding event-order NLLs decrease.
This redistribution is consistent with the price and size dependencies in \cref{tab:field_validity_conditions}, which allow ground-truth reference-order fields to inform subsequent event-order prediction under teacher forcing.
During inference, ReLOBGen selects an eligible resting order from a learned distribution and provides its current attributes before generating the event order, allowing subsequent generation to be conditioned and constrained by the selected reference.
This use of the reference during generation may contribute to ReLOBGen's improved rollout realism.

\vspace{-2mm}
\section{Conclusion}
\label{sec:conclusion}
\vspace{-2mm}

We introduced ReLOBGen, a method for generating LOB messages that are replayable by construction against the current market state.
We organized generation around two replayability conditions---reference eligibility and event-order compatibility---and satisfied them through a reference-first factorization that selects an eligible resting order before constraining event-order generation through inference-time masking.
We further proposed a learned, efficiently computable distribution over eligible resting orders for realistic reference selection.
Empirically, ReLOBGen achieved 100\% replayability in 500-message rollouts on GOOG and INTC, improved most evaluated realism metrics over the LOBS5 baseline, and achieved a $2.7\text{--}3.6\times$ speedup per replayed message.
These results show that replayability by construction can be achieved alongside improved rollout realism and computational efficiency.
Future work can build on ReLOBGen to train and evaluate trading strategies and analyze market impact in closed-loop market simulation.

\bibliography{iclr2027_conference}
\bibliographystyle{iclr2027_conference}

\clearpage
\appendix

\section{Post-processing in prior methods}
\label{app:post-process}

Prior LOB message generators do not guarantee that their raw outputs can be replayed on the current LOB.
Hence, their market-simulation pipelines rely on model-specific, often heuristic, post-processing to convert raw outputs into simulator actions.

The design of this post-processing depends on whether a generated message is treated as an exchange-side record of an event that has already occurred or as a trader-side instruction submitted to the matching engine.
LOBS5 corrects generated messages into exchange-side records, whereas MarS generates trader-side instructions.
MarketGPT incorporates aspects of both designs.

\subsection[LOBS5]{LOBS5~\citep{nagy2023generative}}
\label{app:lobs5}

LOBS5 uses the generated-message field layout $\widehat{M}_t=(\widehat{E}_t,\widehat{X}_t,\widehat{R}_t)$, with the reference fields appearing last.
Here, $\widehat{E}_t$ denotes event type and side, $\widehat{X}_t$ denotes the event order, and $\widehat{R}_t$ denotes the referenced order.

\paragraph{$\widehat E_t[\mathrm{type}]=\mathrm{Add}$.}
\begin{itemize}
    \item \textbf{Rejection:} LOBS5 rejects the message if $\widehat X_t[\mathrm{price}]$ is marketable against the opposite best price.
    \item \textbf{Correction:} Otherwise, LOBS5 replaces any generated non-N/A reference field in $\widehat R_t$ with N/A.
    No other correction is applied.
\end{itemize}

\paragraph{$\widehat E_t[\mathrm{type}]\in\{\mathrm{Cancel},\mathrm{Delete}\}$.}
\begin{itemize}
    \item \textbf{Rejection:} LOBS5 matches a resting order $o\in\OO_{t-1}$ to $\widehat R_t$ under $(\text{price},\text{size},\text{time}) \rightarrow (\text{price},\text{size}) \rightarrow (\text{price})$.
    The final price-only fallback applies only to INIT liquidity, which represents the initial book state at each side--price level as a single order.
    The message is rejected if no such $o$ is found.
    \item \textbf{Correction:} If such an order $o$ is found, LOBS5 sets
    $$ R_t=o,\quad X_t[\mathrm{price}]=R_t[\mathrm{price}],\quad X_t[\mathrm{size}]=\min\{\widehat X_t[\mathrm{size}],R_t[\mathrm{size}]\}. $$
    It sets $E_t[\mathrm{type}]$ to Cancel if $X_t[\mathrm{size}]<R_t[\mathrm{size}]$, and to Delete otherwise.
    \item \textbf{Our reproduction:} We first attempt an exact match against the reconstructed resting-order set, then fall back to full and relaxed matching within the message history, following the original code.
    We use exact resting-order initialization and compare it with INIT-order initialization in \cref{app:init_fallback}.
\end{itemize}

\paragraph{$\widehat E_t[\mathrm{type}]=\mathrm{Execute}$.}
\begin{itemize}
    \item \textbf{Rejection:} The message is rejected if no executable order is found.
    \item \textbf{Correction:} Otherwise, LOBS5 ignores $\widehat R_t$ and sets $R_t=\operatorname{FrontOfQueue}(\OO_{t-1}^{(\widehat E_t[\mathrm{side}])})$.
    It then sets $X_t[\mathrm{price}]=R_t[\mathrm{price}]$ and $X_t[\mathrm{size}]=\min\{\widehat X_t[\mathrm{size}],R_t[\mathrm{size}]\}$.
\end{itemize}

\subsection[MarketGPT]{MarketGPT~\citep{wheeler2024marketgpt}}
\label{app:marketgpt}

MarketGPT uses the generated-message field layout $\widehat{M}_t=(\widehat{E}_t,\widehat{X}_t,\widehat{R}_t)$, with an additional $\widehat X_t[\mathrm{remaining\ size}]$ field and five event types: Add, Execute, Execute-at-different-price, Cancel/Delete, and Replace.
Execute-at-different-price is unsupported in the released generated rollouts, so we omit it below.
For Cancel/Delete, $\widehat X_t[\mathrm{size}]$ and $\widehat X_t[\mathrm{remaining\ size}]$ denote the canceled and remaining quantities, respectively.

\paragraph{$\widehat E_t[\mathrm{type}]=\mathrm{Add}$.}
\begin{itemize}
    \item \textbf{Simulator handling:} The generated message is submitted directly as a limit order without correction or rejection.
    If it is marketable, the matching engine executes it against the opposite book in price--time priority, and any unfilled quantity rests at the generated limit price.
\end{itemize}

\paragraph{$\widehat E_t[\mathrm{type}]=\mathrm{Execute}$.}
\begin{itemize}
    \item \textbf{Simulator handling:} MarketGPT ignores $\widehat R_t$ and $\widehat X_t[\mathrm{price}]$ and treats the generated side and size as a market order.
    It repeatedly matches opposite-side resting orders in price--time priority until the generated size is filled or the opposite book is exhausted, discarding any unfilled quantity.
\end{itemize}

\paragraph{$\widehat E_t[\mathrm{type}]\in\{\mathrm{Cancel},\mathrm{Delete}\}$.}
\begin{itemize}
    \item \textbf{Rejection:} MarketGPT matches a resting order $o\in\OO_{t-1}$ to $(\widehat X_t[\mathrm{price}],\widehat X_t[\mathrm{size}]+\widehat X_t[\mathrm{remaining\ size}],\widehat R_t[\mathrm{time}])$ under $(\text{price},\text{size},\text{time}) \rightarrow (\text{price},\text{size}) \rightarrow (\text{price},\text{time}) \rightarrow (\text{price})$.
    The final price-only fallback selects the front-of-queue resting order at $\widehat X_t[\mathrm{price}]$.
    The message is rejected if no such $o$ is found.
    \item \textbf{Correction:} If such an order $o$ is found, let $\widehat\rho_t=\widehat X_t[\mathrm{size}]/(\widehat X_t[\mathrm{size}]+\widehat X_t[\mathrm{remaining\ size}])$ denote the generated cancellation ratio.
    MarketGPT sets
    $$ R_t=o,\quad X_t[\mathrm{size}]=\left\lfloor R_t[\mathrm{size}]\widehat\rho_t\right\rfloor,\quad X_t[\mathrm{remaining\ size}]=R_t[\mathrm{size}]-X_t[\mathrm{size}], $$
    preserving the generated cancellation ratio for the matched resting order up to integer rounding.
\end{itemize}

\paragraph{$\widehat E_t[\mathrm{type}]=\mathrm{Replace}$.}
\begin{itemize}
    \item \textbf{Rejection:} MarketGPT matches a resting order $o\in\OO_{t-1}$ to $\widehat R_t$ under $(\text{price},\text{size},\text{time}) \rightarrow (\text{price},\text{size}) \rightarrow (\text{price},\text{time}) \rightarrow (\text{price})$.
    The final price-only fallback selects the front-of-queue resting order at $\widehat R_t[\mathrm{price}]$.
    The message is rejected if no such $o$ is found.
    \item \textbf{Correction:} If such an order $o$ is found, MarketGPT sets $R_t=o$ and retains $X_t=\widehat X_t$.
\end{itemize}

\subsection[MarS]{MarS~\citep{li2025mars}}
\label{app:mars}

In our notation, MarS represents each generated output as $\widehat M_t=(\widehat E_t,\widehat X_t)$ without reference-order fields, where $\widehat E_t\in\{\mathrm{Bid},\mathrm{Ask},\mathrm{Cancel}\}$.
It interprets each output as a trader-side instruction to the simulated clearing house rather than as a complete exchange message.

\paragraph{$\widehat E_t\in\{\mathrm{Bid},\mathrm{Ask}\}$.}
\begin{itemize}
    \item \textbf{Simulator handling:} MarS submits the generated Bid or Ask directly as a limit order without correction or rejection.
    If it is marketable, the simulated clearing house executes it against opposite-side resting orders, and any unfilled quantity rests at the generated price.
\end{itemize}

\paragraph{$\widehat E_t=\mathrm{Cancel}$.}
\begin{itemize}
    \item \textbf{Rejection:} If no resting order is available for cancellation, MarS discards the generated cancellation by returning an empty simulator action.
    The discarded output is not added to the model's order history, and the agent immediately samples again without rolling back the elapsed simulation time.
    \item \textbf{Correction:} MarS projects the generated cancellation onto the current resting-order set $\OO_{t-1}$ by selecting an available resting order $o\in\OO_{t-1}$ whose price $o[\mathrm{price}]$ is closest to $\widehat X_t[\mathrm{price}]$.
    It assigns actual order IDs and sides and distributes $\widehat X_t[\mathrm{size}]$ across the resting orders at $o[\mathrm{price}]$, capped by their available size.
\end{itemize}

\clearpage

\section{Experimental details}
\label{app:experimental_details}

\subsection{Preprocessing}
\label{app:data_preparation}
We convert XNAS.ITCH MBO data for GOOG and INTC from Databento\footnote{\url{https://databento.com/portal/catalog/us-equities\#XNAS.ITCH}} into LOBSTER-compatible messages~\citep{Huang2011LOBSTER} and follow the preprocessing pipeline of LOBS5~\citep{nagy2023generative}, except for the reference-order representation.
See below for details.

\paragraph{Databento MBO to LOBSTER-compatible message.}
Within each stream identified by \texttt{(publisher\_id, instrument\_id, channel\_id)}, consecutive records in feed order with the same \texttt{sequence} value form an event group, which is converted according to \cref{tab:mbo_action_conversion}.
Under these conversion rules, individual fields are mapped as shown in \cref{tab:mbo_lobster_field_correspondence}, with format conversions applied as needed.

\begin{table}[H]
  \centering
  \caption{\textbf{Conversion of Databento MBO action patterns to LOBSTER-compatible messages.}
    Reset clears the order state once at session start without emitting a message.
    Sole trades (single-\texttt{T} groups with \texttt{side=N}) may correspond to Type 5 but are omitted because their side is unknown.}
  \label{tab:mbo_action_conversion}
  \renewcommand{\arraystretch}{1.1}
  \setlength{\tabcolsep}{5pt}
  \scalebox{0.95}{%
    \begin{tabular}{lll}
      \toprule
      MBO action pattern & LOBSTER output & Side source \\
      \midrule
      Add (\texttt{A}) & Type 1: Add & side of \texttt{A} \\
      Cancel (\texttt{C}) & Type 2/3: Cancel/Delete & side of \texttt{C} \\
      Replace (\texttt{C} $\rightarrow$ \texttt{A}) & Type 2/3 $\rightarrow$ Type 1 & side of each \texttt{C} / \texttt{A} \\
      Visible execution (\texttt{T} $\rightarrow$ \texttt{F} $\rightarrow$ \texttt{C}) & Type 4: Execute & side of \texttt{F} \\
      Cross trade (\texttt{F} $\rightarrow$ \texttt{C}) & Type 6: Cross trade & side of \texttt{C} \\
      Reset (\texttt{R}) & No LOBSTER message & --- \\
      Sole trade (\texttt{T} with \texttt{side=N}) & No LOBSTER message & --- \\
      \bottomrule
    \end{tabular}
  }
\end{table}

\begin{table}[H]
  \centering
  \caption{\textbf{Field correspondence between Databento MBO and LOBSTER-compatible messages.}}
  \label{tab:mbo_lobster_field_correspondence}
  \renewcommand{\arraystretch}{1.1}
  \setlength{\tabcolsep}{5pt}
  \scalebox{0.95}{%
    \begin{tabular}{lll}
      \toprule
      MBO field & LOBSTER-compatible field & Description \\
      \midrule
      \texttt{ts\_event} & \texttt{Time} & Exchange-recorded event timestamp \\
      \texttt{action} & \texttt{Type} & Order-book event category \\
      \texttt{order\_id} & \texttt{Order ID} & Individual order identifier \\
      \texttt{side} & \texttt{Direction} & Buy or sell side \\
      \texttt{price} & \texttt{Price} & Order or execution price \\
      \texttt{size} & \texttt{Size} & Order, cancellation, or execution quantity \\
      \bottomrule
    \end{tabular}
  }
\end{table}

\paragraph{Message representation.}
\cref{tab:normalization_encoding} summarizes normalization and encoding in ReLOBGen.
All fields share a single vocabulary of 12,011 tokens, including three special tokens for masking, hiding, and not-applicable values.
The event order's absolute timestamp is determined by adding its inter-arrival time to the previous event's timestamp, rather than generated by the model.

LOBS5~\citep{nagy2023generative} copies the encoded field values from the referenced order's original Add message into the reference-order fields, retaining the price encoding relative to the mid-price at submission rather than the current mid-price.

\begin{table}[H]
  \centering
  \caption{\textbf{Normalization and encoding in ReLOBGen.}
    Prices are encoded as tick offsets from the tick-aligned mid-price, clipped to $[-999,999]$: $\Delta p=\operatorname{clip}((p_{\mathrm{order}}-p_{\mathrm{mid}})/\tau,-999,999)$.
    Here, $p_{\mathrm{mid}}=\tau\lfloor(p_{\mathrm{ask},1}+p_{\mathrm{bid},1})/(2\tau)\rfloor$, $p_{\mathrm{ask},1}$ and $p_{\mathrm{bid},1}$ denote the best ask and bid prices, respectively, and $\tau=100$.
    Each time component is encoded in base 1000, with one token per three-digit group.}
  \label{tab:normalization_encoding}
  \renewcommand{\arraystretch}{1.1}
  \setlength{\tabcolsep}{8pt}
  \scalebox{0.95}{%
    \begin{tabular}{lcl}
      \toprule
      Field & \# Tokens & Normalization and encoding \\
      \midrule
      Event type & 1 & No normalization \\
      Event side & 1 & Ask: 0 / Bid: 1 \\
      \midrule
      Reference-order price & 2 & $(\operatorname{sign}(\Delta p), |\Delta p|)$ \\
      Reference-order size & 1 & $\min(\text{remaining size}, 9999)$ \\
      Reference-order time~(s) & 2 & Seconds component \\
      Reference-order time~(ns) & 3 & Nanoseconds component \\
      \midrule
      Event-order price & 2 & $(\operatorname{sign}(\Delta p), |\Delta p|)$ \\
      Event-order size & 1 & $\min(\text{size}, 9999)$ \\
      Event-order $\Delta t$~(s) & 1 & Seconds component \\
      Event-order $\Delta t$~(ns) & 3 & Nanoseconds component \\
      Event-order time~(s) & 2 & Seconds component \\
      Event-order time~(ns) & 3 & Nanoseconds component \\
      \bottomrule
    \end{tabular}
  }
\end{table}

\paragraph{Book representation.}
We reconstruct book states by replaying the message sequence.
Then, following LOBS5~\citep{nagy2023generative}, we represent each book state as a 501-dimensional continuous vector comprising the mid-price change in ticks and signed volumes on a 500-slot price grid centered on the tick-aligned mid-price:
\begin{align}
\mathbf b_t &= \left(\Delta p_{\mathrm{mid},t}/\tau, x_{t,0},\ldots,x_{t,499}\right)\in\mathbb R^{501},
\end{align}
where $\Delta p_{\mathrm{mid},t}=p_{\mathrm{mid},t}-p_{\mathrm{mid},t-1}$ and $\tau=100$.
We map the top 10 ask and bid levels onto this grid by their tick offsets from the mid-price.
Each slot stores the total quantity divided by 1000, negative for asks and positive for bids.
Levels outside the grid are omitted, and empty slots are zero.

\paragraph{Filtering.}
For training, we retain only Type 1 (Add), Type 2 (Cancel), Type 3 (Delete), and Type 4 (Execute) messages whose prices lie between the 10th-best bid and 10th-best ask prices at the time of the event.
We use only 500-message sequences fully contained within regular trading hours~(09:30--16:00 ET).

\subsection{Training configuration}
\label{app:model_training}

We follow the scaled-up LOBS5 architecture used in LOB-Bench~\citep{nagy2023generative,nagy2025lob} and likewise use one year of training data for both stocks, covering Jan.~1--Dec.~31, 2025, with Jan.~1--15, 2026 for validation.
We train the message generation model $\pi_\theta$ first, followed by the query and order projection heads $f_{\mathrm{query}}$ and $f_{\mathrm{order}}$ on the same data periods.
\cref{tab:training_configuration} summarizes the training settings, and \cref{tab:model_architecture} details the architecture of each component.
Training budgets count message windows for $\pi_\theta$ and window--mid-price pairs for the projection heads, with 256 mid-prices per window.


\begin{table}[H]
  \centering
  \caption{\textbf{Training configuration.}
    In both training stages, learning rates use linear warmup over the first 10\% of training steps, followed by cosine decay to 10\% of their peak values.}
  \label{tab:training_configuration}
  \renewcommand{\arraystretch}{1.1}
  \setlength{\tabcolsep}{3.5pt}
  \scalebox{0.95}{%
    \begin{tabular}{ccccccc}
      \toprule
      Training stage & Dataset & Budget & Batch size & Optimizer & Peak LR & GPU hours \\
      \midrule
      \multirow{2}{*}{Base model} & GOOG & 100M & 512 & \multirow{2}{*}{Adam} & $5\times10^{-4}$ & 8 L40S $\times$ 54 h \\
      & INTC & 50M & 512 & & $5\times10^{-4}$ & 8 L40S $\times$ 27 h \\
      \midrule
      \multirow{2}{*}{Reference selection} & GOOG & 200M & 256 & \multirow{2}{*}{AdamW} & $1\times10^{-4}$ & 1 A6000 $\times$ 33 h \\
      & INTC & 100M & 256 & & $1\times10^{-4}$ & 1 A6000 $\times$ 25 h \\
      \bottomrule
    \end{tabular}
  }
\end{table}

\begin{table}[H]
  \centering
  \caption{\textbf{Model architecture.}
    $\pi_\theta$ is the message generation model; $f_{\mathrm{query}}$ and $f_{\mathrm{order}}$ are the query and order projection heads, respectively.}
  \label{tab:model_architecture}
  \renewcommand{\arraystretch}{1.1}
  \setlength{\tabcolsep}{5pt}
  \scalebox{0.95}{%
    \begin{tabular}{ccl}
      \toprule
      Model & \# Params & Details \\
      \midrule
      $\pi_\theta$ & 35.78M &
      \begin{tabular}{@{}l@{}}
        \textbullet\ S5 backbone: $d_{\mathrm{model}}=512$; SSM state size 512. \\
        \textbullet\ 2 S5 message-encoder layers; 1 pre- and 1 post-book-encoder layer. \\
        \textbullet\ 12 S5 layers for message--book fusion.
      \end{tabular} \\
      \midrule
      $f_{\mathrm{query}}$ & 0.13M &
      \begin{tabular}{@{}l@{}}
        \textbullet\ Layer-normalized 512-d state concatenated with a GELU-projected \\
        mid-price displacement ($1\rightarrow512$). \\
        \textbullet\ $1024\rightarrow128$ projection followed by layer normalization.
      \end{tabular} \\
      \midrule
      $f_{\mathrm{order}}$ & 2.16M &
      \begin{tabular}{@{}l@{}}
        \textbullet\ Eight 512-d token embeddings flattened into a \\
        $4096\rightarrow512\rightarrow128$ MLP, with GELU after the first layer \\
        and output layer normalization.
      \end{tabular} \\
      \bottomrule
    \end{tabular}
  }
\end{table}



\subsection{Rollout protocol}
\label{app:rollout_evaluation}

For each stock, we use 100 rollout windows from each of 10 test days, January 16--30, 2026.
Each rollout starts from the exact resting-order set within the top 20 price levels on each side, providing a buffer of deeper resting orders that may enter the top 10 price levels during the rollout.
Consistent with training, model LOB-state inputs and eligible reference sets are restricted to the current top 10 price levels on each side.
We do not apply temperature scaling or top-$k$/top-$p$ filtering during either resting-order selection or token sampling.
To avoid spending excessive computation on stalled rollouts, we restart a rollout with a different random seed after 100 consecutive failed attempts.

\clearpage
\section{Additional experimental results}
\label{app:additional_results}

\subsection{Event-wise replayability}
\label{app:eventwise_replayability}

\cref{tab:eventwise_replayability} reports post-processing and replayability violation rates by event type on GOOG and INTC.
Both LOBS5 baselines rarely require post-processing for add events, but frequently reject cancel and delete events and correct all execute events.
Across non-add event types, reference violations are more frequent than event-order violations on both stocks.
ReLOBGen records no violations and requires no post-processing for any event type.


\begin{table}[H]
  \centering
  \caption{\textbf{Event-wise post-processing rates and replayability violations.}
  Positive rates below $0.1\%$ are reported as \textless0.1\%.
  Lower is better, and bold indicates the lowest value.}
  \label{tab:eventwise_replayability}
  \renewcommand{\arraystretch}{1.1}
  \setlength{\tabcolsep}{3.5pt}
  \scalebox{0.90}{%
    \begin{tabular}{@{}cllrrrrr@{}}
      \toprule
      & & & & \multicolumn{2}{c}{Post-processing} & \multicolumn{2}{c}{Replayability violations} \\
      \cmidrule(lr){5-6}\cmidrule(lr){7-8}
      Ticker & Event & Method & \multicolumn{1}{c}{Attempts} & \multicolumn{1}{c}{Correction} & \multicolumn{1}{c}{Rejection} & \multicolumn{1}{c}{Reference} & \multicolumn{1}{c}{Event-order} \\
      \midrule
      \multirow{12}{*}{GOOG}
      & \multirow{3}{*}{Add} & LOBS5~(ref.-last) & 255,720 & \textbf{0.0\%} & 0.2\% & \textbf{0.0\%} & 0.2\% \\
      & & LOBS5~(ref.-first) & 259,199 & \textbf{0.0\%} & \textless0.1\% & \textbf{0.0\%} & \textless0.1\% \\
      & & \cellcolor{gray!10}ReLOBGen & \cellcolor{gray!10}243,723 & \cellcolor{gray!10}\textbf{0.0\%} & \cellcolor{gray!10}\textbf{0.0\%} & \cellcolor{gray!10}\textbf{0.0\%} & \cellcolor{gray!10}\textbf{0.0\%} \\
      \cmidrule(lr){2-8}
      & \multirow{3}{*}{Cancel} & LOBS5~(ref.-last) & 4,692 & 9.9\% & 67.6\% & 76.1\% & 5.1\% \\
      & & LOBS5~(ref.-first) & 2,781 & 9.4\% & 67.7\% & 76.7\% & 1.8\% \\
      & & \cellcolor{gray!10}ReLOBGen & \cellcolor{gray!10}1,058 & \cellcolor{gray!10}\textbf{0.0\%} & \cellcolor{gray!10}\textbf{0.0\%} & \cellcolor{gray!10}\textbf{0.0\%} & \cellcolor{gray!10}\textbf{0.0\%} \\
      \cmidrule(lr){2-8}
      & \multirow{3}{*}{Delete} & LOBS5~(ref.-last) & 353,723 & 8.9\% & 47.8\% & 56.7\% & \textless0.1\% \\
      & & LOBS5~(ref.-first) & 379,291 & 8.0\% & 50.1\% & 58.1\% & \textless0.1\% \\
      & & \cellcolor{gray!10}ReLOBGen & \cellcolor{gray!10}209,395 & \cellcolor{gray!10}\textbf{0.0\%} & \cellcolor{gray!10}\textbf{0.0\%} & \cellcolor{gray!10}\textbf{0.0\%} & \cellcolor{gray!10}\textbf{0.0\%} \\
      \cmidrule(lr){2-8}
      & \multirow{3}{*}{Execute} & LOBS5~(ref.-last) & 58,814 & 100.0\% & \textbf{0.0\%} & 100.0\% & 3.1\% \\
      & & LOBS5~(ref.-first) & 50,781 & 100.0\% & \textbf{0.0\%} & 100.0\% & 0.9\% \\
      & & \cellcolor{gray!10}ReLOBGen & \cellcolor{gray!10}45,824 & \cellcolor{gray!10}\textbf{0.0\%} & \cellcolor{gray!10}\textbf{0.0\%} & \cellcolor{gray!10}\textbf{0.0\%} & \cellcolor{gray!10}\textbf{0.0\%} \\
      \midrule
      \multirow{12}{*}{INTC}
      & \multirow{3}{*}{Add} & LOBS5~(ref.-last) & 264,173 & \textless0.1\% & 0.7\% & \textless0.1\% & 0.7\% \\
      & & LOBS5~(ref.-first) & 269,847 & \textbf{0.0\%} & 0.2\% & \textbf{0.0\%} & 0.2\% \\
      & & \cellcolor{gray!10}ReLOBGen & \cellcolor{gray!10}253,908 & \cellcolor{gray!10}\textbf{0.0\%} & \cellcolor{gray!10}\textbf{0.0\%} & \cellcolor{gray!10}\textbf{0.0\%} & \cellcolor{gray!10}\textbf{0.0\%} \\
      \cmidrule(lr){2-8}
      & \multirow{3}{*}{Cancel} & LOBS5~(ref.-last) & 1,494 & 10.8\% & 66.9\% & 73.8\% & 13.9\% \\
      & & LOBS5~(ref.-first) & 1,377 & 19.0\% & 50.5\% & 65.4\% & 13.8\% \\
      & & \cellcolor{gray!10}ReLOBGen & \cellcolor{gray!10}1,068 & \cellcolor{gray!10}\textbf{0.0\%} & \cellcolor{gray!10}\textbf{0.0\%} & \cellcolor{gray!10}\textbf{0.0\%} & \cellcolor{gray!10}\textbf{0.0\%} \\
      \cmidrule(lr){2-8}
      & \multirow{3}{*}{Delete} & LOBS5~(ref.-last) & 311,405 & 12.1\% & 30.3\% & 42.4\% & 0.1\% \\
      & & LOBS5~(ref.-first) & 291,193 & 18.0\% & 25.6\% & 43.6\% & \textless0.1\% \\
      & & \cellcolor{gray!10}ReLOBGen & \cellcolor{gray!10}223,952 & \cellcolor{gray!10}\textbf{0.0\%} & \cellcolor{gray!10}\textbf{0.0\%} & \cellcolor{gray!10}\textbf{0.0\%} & \cellcolor{gray!10}\textbf{0.0\%} \\
      \cmidrule(lr){2-8}
      & \multirow{3}{*}{Execute} & LOBS5~(ref.-last) & 19,952 & 100.0\% & \textbf{0.0\%} & 100.0\% & 3.4\% \\
      & & LOBS5~(ref.-first) & 13,473 & 100.0\% & \textbf{0.0\%} & 100.0\% & 2.3\% \\
      & & \cellcolor{gray!10}ReLOBGen & \cellcolor{gray!10}21,072 & \cellcolor{gray!10}\textbf{0.0\%} & \cellcolor{gray!10}\textbf{0.0\%} & \cellcolor{gray!10}\textbf{0.0\%} & \cellcolor{gray!10}\textbf{0.0\%} \\
      \bottomrule
    \end{tabular}
  }
\end{table}

\clearpage
\subsection{Effect of resting-order initialization on replayability}
\label{app:init_fallback}

We compare two ways of initializing the resting-order set at the start of a rollout.
INIT-order initialization, as used in the original LOBS5 implementation, represents the total volume at each side--price level of the initial LOB state as a single INIT order, whereas exact resting-order initialization preserves the individual orders reconstructed from market-by-order data.
INIT orders do not correspond to individual orders in the model's training messages and can only be matched to generated references through price-only matching during post-processing.
Our evaluation uses exact resting-order initialization because the exact resting-order set is available from our data.

\begin{figure}[H]
    \centering
    \includegraphics[width=\textwidth]{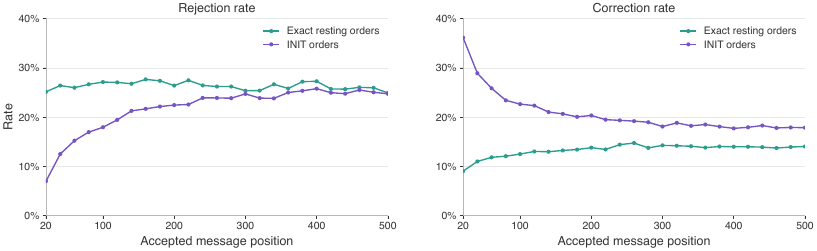}
    \caption{\textbf{Effect of resting-order initialization on replayability.}
    We report rejection and correction rates for LOBS5~(ref.-last) on GOOG under INIT-order and exact resting-order initialization.
    Rates are computed over generation attempts grouped into 20-message bins by accepted-message position.}
    \label{fig:init_fallback}
\end{figure}

\cref{fig:init_fallback} shows that LOBS5~(ref.-last) still requires correction and rejection under INIT-order initialization on GOOG.
With INIT orders, rejection rates start lower but approach those under exact initialization by the end of the rollout.
Together with the higher correction rates, this pattern suggests that the initial reduction in rejection reflects the looser matching available for INIT orders, rather than improved replayability of the raw generated messages.

\clearpage
\subsection{Additional LOB-Bench results}
\label{app:additional_lob_bench}

\begin{figure}[H]
\centering
\includegraphics[width=\textwidth]{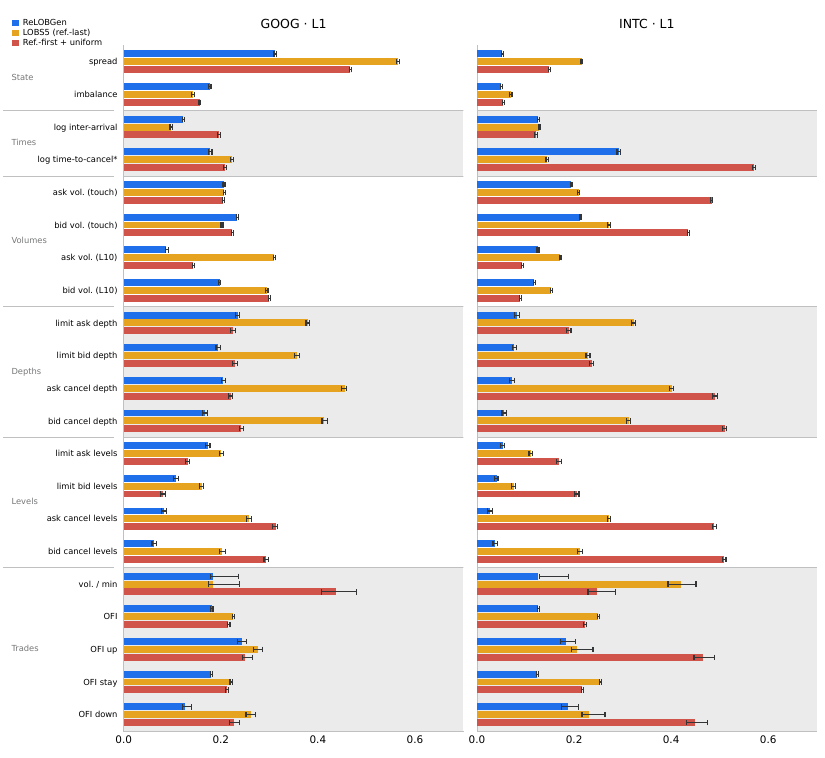}
\caption{\textbf{LOB-Bench evaluation: unconditional $L_1$ distances.}
Log time-to-cancel* measures time-to-cancel for all orders canceled during the rollout using exact resting-order information, including cancellations of orders present in the initial LOB that were excluded from the original LOB-Bench implementation.
Error bars indicate 99\% percentile bootstrap confidence intervals.
Lower is better.}
\label{fig:lob-bench-uncond}
\end{figure}

\begin{figure}[H]
\centering
\includegraphics[width=\textwidth]{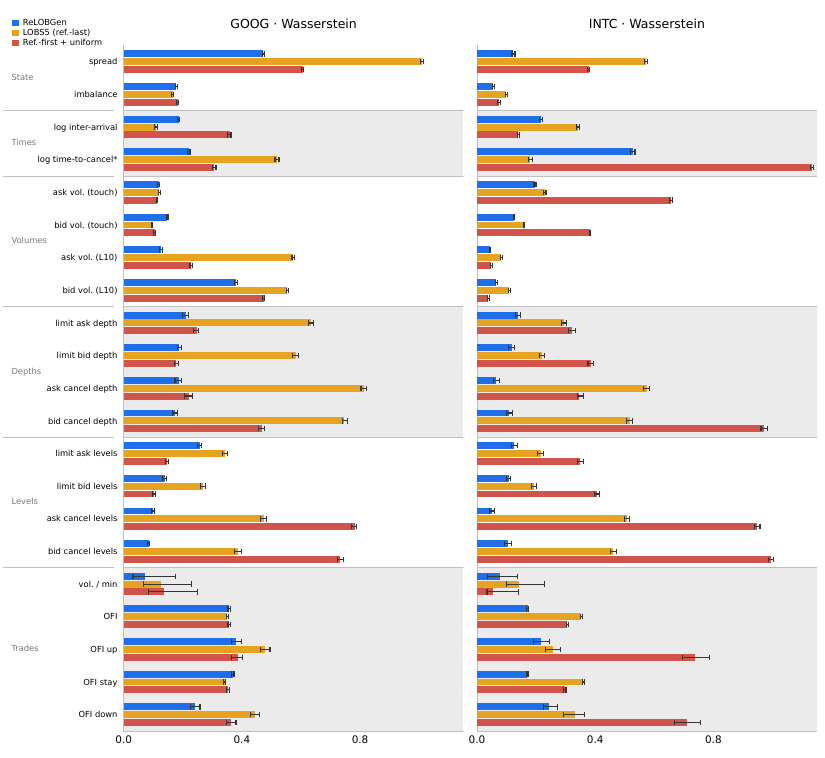}
\caption{\textbf{LOB-Bench evaluation: unconditional Wasserstein distances.}
Log time-to-cancel* measures time-to-cancel for all orders canceled during the rollout using exact resting-order information, including cancellations of orders present in the initial LOB that were excluded from the original LOB-Bench implementation.
Error bars indicate 99\% percentile bootstrap confidence intervals.
Lower is better.}
\label{fig:lob_bench_unconditional_w1}
\end{figure}

\paragraph{Unconditional distributions.}
ReLOBGen achieves broadly lower unconditional $L_1$ and Wasserstein distances than LOBS5 across metric groups~(\cref{tab:lob_bench_unconditional_summary_l1,tab:lob_bench_unconditional_summary_w1}).
At the individual-metric level, ReLOBGen achieves lower Wasserstein point estimates on 16 of 21 metrics for GOOG, with statistically significant improvements on 14 of those 16.
On INTC, it achieves lower point estimates on 20 of 21 metrics, with statistically significant improvements on 18 of those 20, using the same CI non-overlap criterion as in \cref{subsec:rollout_realism}.
\cref{fig:lob-bench-uncond,fig:lob_bench_unconditional_w1} show the individual metric distances and their 99\% bootstrap confidence intervals for $L_1$ and Wasserstein distances, respectively.


\begin{table}[H]
  \centering
  \caption{\textbf{LOB-Bench evaluation: metric-group summary of unconditional Wasserstein distances.}
  Overall averages all metrics, while each metric-group column averages the metrics in its group.
  Lower is better, and bold indicates the lowest value.}
  \label{tab:lob_bench_unconditional_summary_w1}
  \renewcommand{\arraystretch}{1.1}
  \setlength{\tabcolsep}{3.5pt}
  \scalebox{0.95}{%
    \begin{tabular}{@{}cl@{\hspace{5pt}}ccccccc@{}}
      \toprule
      & & & \multicolumn{6}{c}{Metric group} \\
      \cmidrule(lr){4-9}
      Ticker & Method & Overall & State & Times & Volumes & Depths & Levels & Trades \\
      \midrule
      \multirow{3}{*}{GOOG}
      & LOBS5~(ref.-last) & 0.43 & 0.59 & 0.31 & 0.34 & 0.69 & 0.37 & 0.35 \\
      & Ref.-first + uniform & 0.33 & 0.39 & 0.33 & 0.23 & 0.28 & 0.44 & 0.32 \\
      & \cellcolor{gray!10}ReLOBGen & \cellcolor{gray!10}\textbf{0.22} & \cellcolor{gray!10}\textbf{0.33} & \cellcolor{gray!10}\textbf{0.20} & \cellcolor{gray!10}\textbf{0.19} & \cellcolor{gray!10}\textbf{0.19} & \cellcolor{gray!10}\textbf{0.14} & \cellcolor{gray!10}\textbf{0.28} \\
      \midrule
      \multirow{3}{*}{INTC}
      & LOBS5~(ref.-last) & 0.30 & 0.34 & \textbf{0.26} & 0.14 & 0.40 & 0.34 & 0.29 \\
      & Ref.-first + uniform & 0.46 & 0.23 & 0.64 & 0.28 & 0.51 & 0.67 & 0.42 \\
      & \cellcolor{gray!10}ReLOBGen & \cellcolor{gray!10}\textbf{0.15} & \cellcolor{gray!10}\textbf{0.09} & \cellcolor{gray!10}0.37 & \cellcolor{gray!10}\textbf{0.11} & \cellcolor{gray!10}\textbf{0.11} & \cellcolor{gray!10}\textbf{0.10} & \cellcolor{gray!10}\textbf{0.18} \\
      \bottomrule
    \end{tabular}
  }
\end{table}

\clearpage

\clearpage

\clearpage
\paragraph{Conditional distributions.}
ReLOBGen achieves lower conditional Wasserstein point estimates than LOBS5 on 5 of 6 metrics across the two stocks~(\cref{tab:lob_bench_cond_impact_w1}).

\begin{table}[H]
  \centering
  \caption{\textbf{LOB-Bench evaluation: conditional Wasserstein distances.}
  Conditional 99\% percentile bootstrap confidence intervals have half-widths below 0.012.
  Lower is better, and bold indicates the lowest value.}
  \label{tab:lob_bench_cond_impact_w1}
  \renewcommand{\arraystretch}{1.1}
  \setlength{\tabcolsep}{3pt}
  \scalebox{0.80}{
    \begin{tabular}{@{}llccc@{}}
      \toprule
      & & \multicolumn{3}{c}{Conditional ($W_1$)} \\
      \cmidrule(lr){3-5}
      Ticker & Method
      & ask vol. $\mid$ spread & spread $\mid$ time & spread $\mid$ volatility \\
      \midrule
      \multirow{3}{*}{GOOG} & LOBS5~(ref.-last)
      & \textbf{0.12} & 1.17 & 0.91 \\
       & Ref.-first + uniform
      & \textbf{0.12} & 0.78 & 1.03 \\
       & \cellcolor{gray!10}ReLOBGen
      & \cellcolor{gray!10}0.17 & \cellcolor{gray!10}\textbf{0.60} & \cellcolor{gray!10}\textbf{0.83} \\
      \midrule
      \multirow{3}{*}{INTC} & LOBS5~(ref.-last)
      & 0.24 & 0.57 & 0.50 \\
       & Ref.-first + uniform
      & 0.67 & 0.36 & 0.23 \\
       & \cellcolor{gray!10}ReLOBGen
      & \cellcolor{gray!10}\textbf{0.21} & \cellcolor{gray!10}\textbf{0.12} & \cellcolor{gray!10}\textbf{0.17} \\
      \bottomrule
    \end{tabular}
  }
\end{table}

\begin{figure}[H]
\centering
\includegraphics[width=0.38\textwidth]{assets/results/lob-bench-impact-goog-response/legend.pdf}

\setlength{\tabcolsep}{1pt}
\renewcommand{\arraystretch}{0.75}
\begin{tabular}{@{}ccc@{}}
\multicolumn{3}{c}{GOOG} \\
\includegraphics[width=0.32\textwidth]{assets/results/lob-bench-impact-goog-response/MO_0.pdf} &
\includegraphics[width=0.32\textwidth]{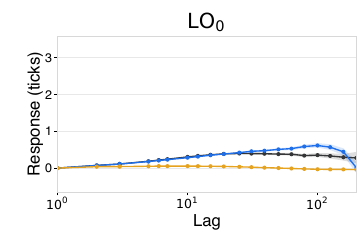} &
\includegraphics[width=0.32\textwidth]{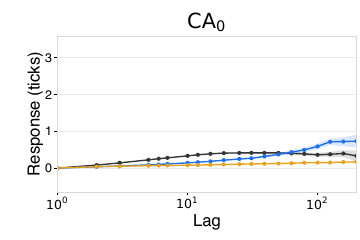} \\
\includegraphics[width=0.32\textwidth]{assets/results/lob-bench-impact-goog-response/MO_1.pdf} &
\includegraphics[width=0.32\textwidth]{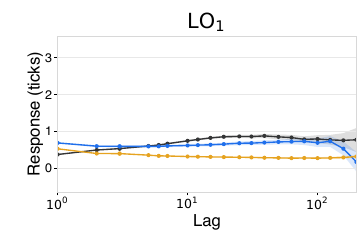} &
\includegraphics[width=0.32\textwidth]{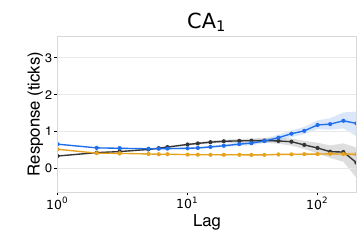} \\[6pt]
\multicolumn{3}{c}{INTC} \\
\includegraphics[width=0.32\textwidth]{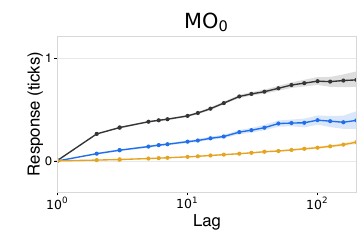} &
\includegraphics[width=0.32\textwidth]{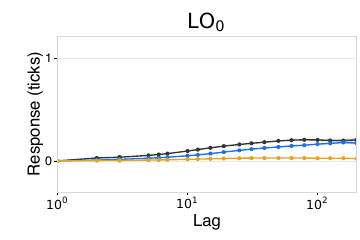} &
\includegraphics[width=0.32\textwidth]{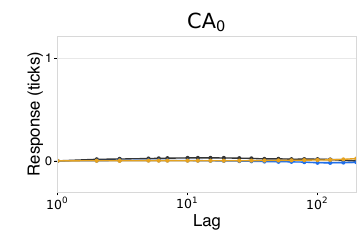} \\
\includegraphics[width=0.32\textwidth]{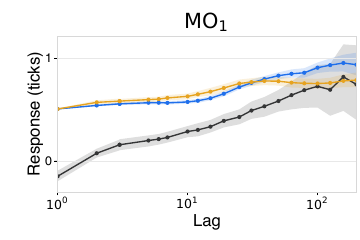} &
\includegraphics[width=0.32\textwidth]{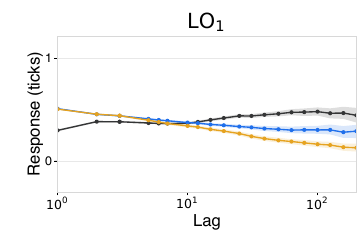} &
\includegraphics[width=0.32\textwidth]{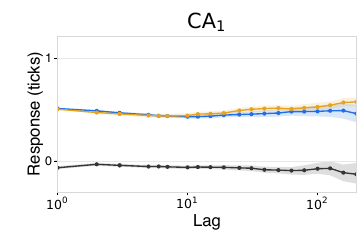}
\end{tabular}
\caption{\textbf{LOB-Bench evaluation: market-impact response curves.}
Responses are plotted against event lag for market orders~(MO), limit orders~(LO), and cancellations~(CA), with subscripts 0 and 1 for events without and with an immediate mid-price change.
Shaded regions indicate 99\% bootstrap confidence intervals.}
\label{fig:lob_bench_impact_response}
\end{figure}

\clearpage
\paragraph{Market-impact responses.}
\cref{fig:lob_bench_impact_response} shows price-response curves following market orders, limit orders, and cancellations on GOOG and INTC.
LOBS5 produces largely flat response curves across all six event types on GOOG and across the three event types without an immediate mid-price change on INTC.
ReLOBGen more closely matches the real responses, with lower market-impact discrepancies than LOBS5 on 9 of 12 metrics across the two stocks~(\cref{tab:lob_bench_cond_impact}).

\begin{figure}[H]
\centering
\includegraphics[width=0.32\textwidth]{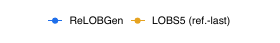}

\setlength{\tabcolsep}{1pt}
\renewcommand{\arraystretch}{0.65}
\begin{tabular}{@{}cc@{\hspace{5pt}}cc@{}}
\multicolumn{2}{c}{GOOG} & \multicolumn{2}{c}{INTC} \\
\includegraphics[width=0.210\textwidth]{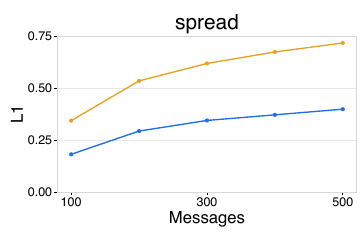} &
\includegraphics[width=0.210\textwidth]{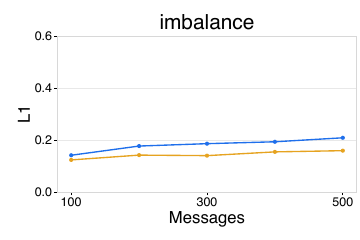} &
\includegraphics[width=0.210\textwidth]{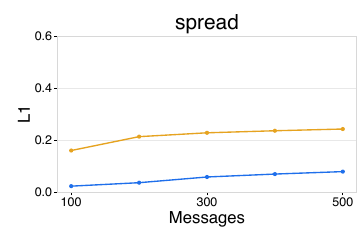} &
\includegraphics[width=0.210\textwidth]{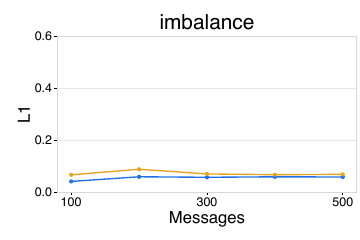} \\
\includegraphics[width=0.210\textwidth]{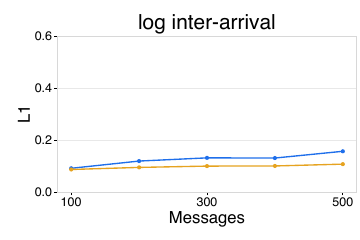} &
\includegraphics[width=0.210\textwidth]{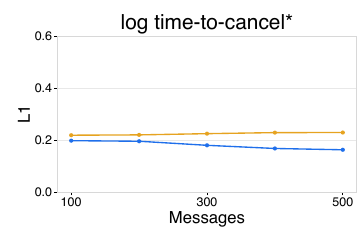} &
\includegraphics[width=0.210\textwidth]{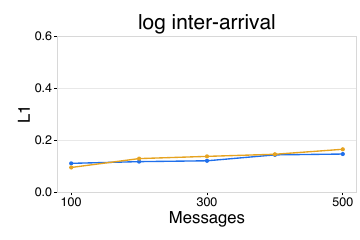} &
\includegraphics[width=0.210\textwidth]{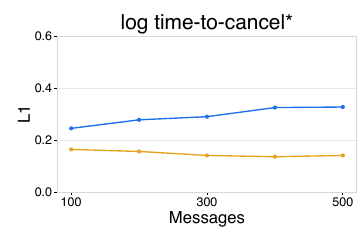} \\
\includegraphics[width=0.210\textwidth]{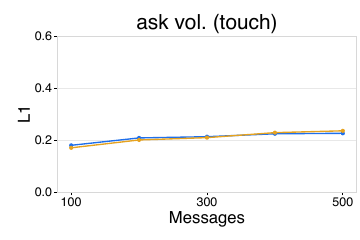} &
\includegraphics[width=0.210\textwidth]{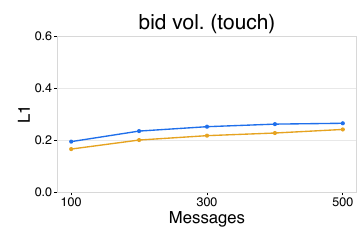} &
\includegraphics[width=0.210\textwidth]{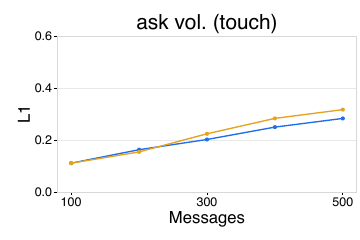} &
\includegraphics[width=0.210\textwidth]{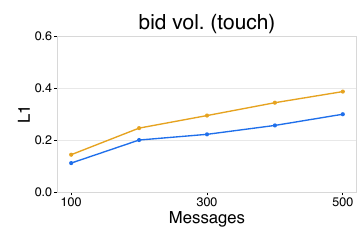} \\
\includegraphics[width=0.210\textwidth]{assets/results/lob-bench-divergence-goog-l1/ask_volume.pdf} &
\includegraphics[width=0.210\textwidth]{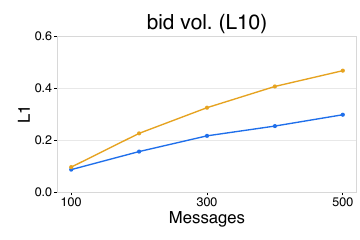} &
\includegraphics[width=0.210\textwidth]{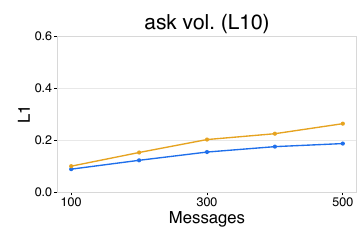} &
\includegraphics[width=0.210\textwidth]{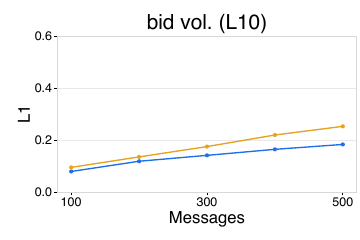} \\
\includegraphics[width=0.210\textwidth]{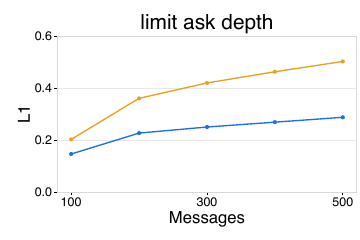} &
\includegraphics[width=0.210\textwidth]{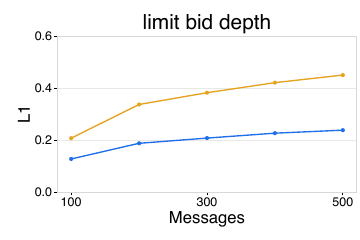} &
\includegraphics[width=0.210\textwidth]{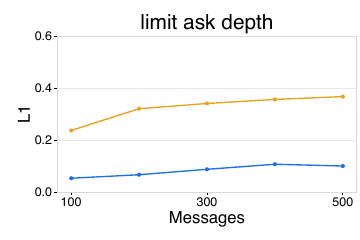} &
\includegraphics[width=0.210\textwidth]{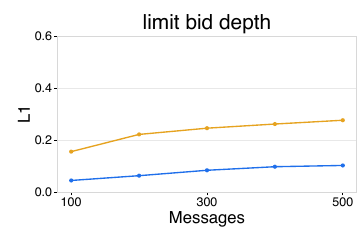} \\
\includegraphics[width=0.210\textwidth]{assets/results/lob-bench-divergence-goog-l1/ask_cancellation_depth.pdf} &
\includegraphics[width=0.210\textwidth]{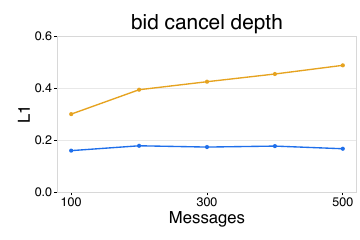} &
\includegraphics[width=0.210\textwidth]{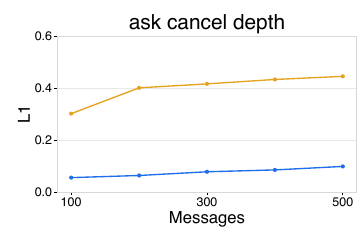} &
\includegraphics[width=0.210\textwidth]{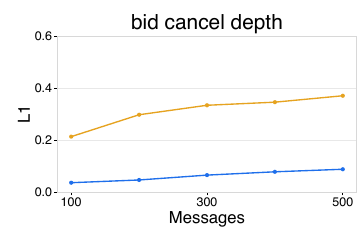} \\
\includegraphics[width=0.210\textwidth]{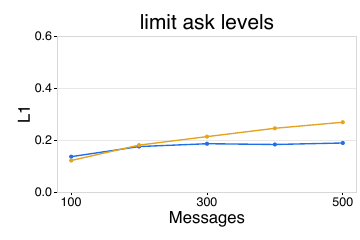} &
\includegraphics[width=0.210\textwidth]{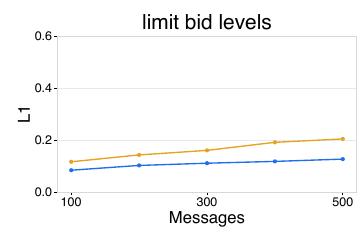} &
\includegraphics[width=0.210\textwidth]{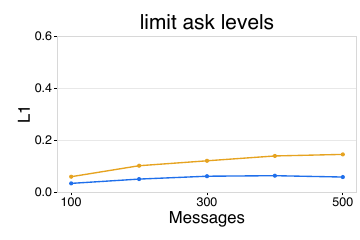} &
\includegraphics[width=0.210\textwidth]{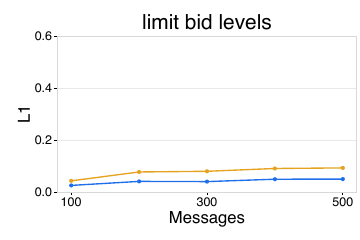} \\
\includegraphics[width=0.210\textwidth]{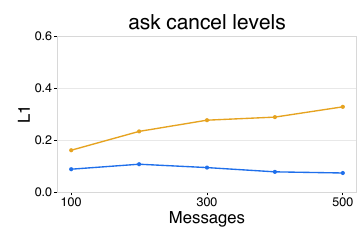} &
\includegraphics[width=0.210\textwidth]{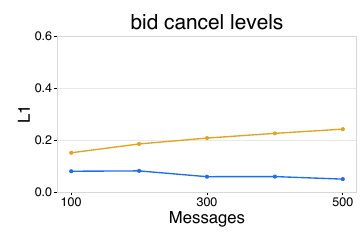} &
\includegraphics[width=0.210\textwidth]{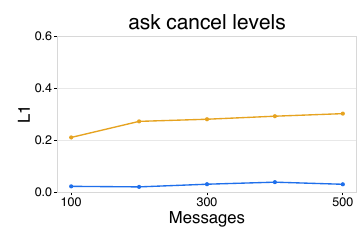} &
\includegraphics[width=0.210\textwidth]{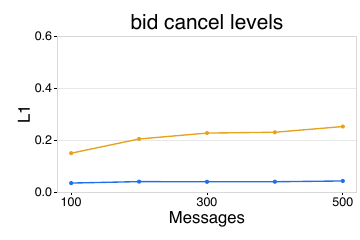} \\
\includegraphics[width=0.210\textwidth]{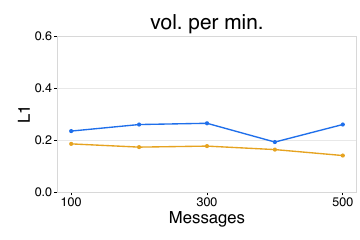} &
& \includegraphics[width=0.210\textwidth]{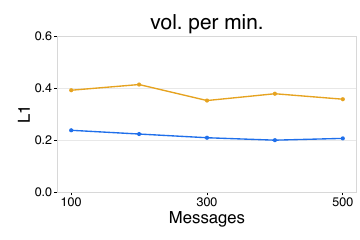} &
\end{tabular}
\caption{\textbf{LOB-Bench evaluation: error accumulation in $L_1$ distances.}
Log time-to-cancel* measures time-to-cancel for all orders canceled during the rollout using exact resting-order information, including cancellations of orders present in the initial LOB that were excluded from the original LOB-Bench implementation.
Lower is better.}
\label{fig:lob_bench_divergence_l1}
\end{figure}

\begin{figure}[H]
\centering
\includegraphics[width=0.32\textwidth]{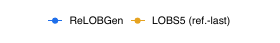}

\setlength{\tabcolsep}{1pt}
\renewcommand{\arraystretch}{0.65}
\begin{tabular}{@{}cc@{\hspace{5pt}}cc@{}}
\multicolumn{2}{c}{GOOG} & \multicolumn{2}{c}{INTC} \\
\includegraphics[width=0.210\textwidth]{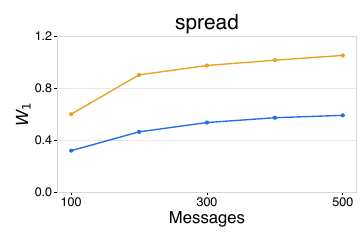} &
\includegraphics[width=0.210\textwidth]{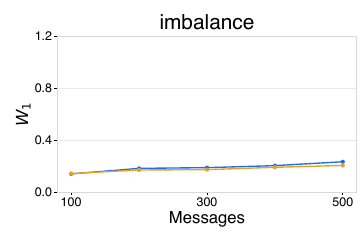} &
\includegraphics[width=0.210\textwidth]{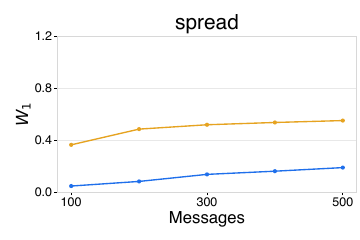} &
\includegraphics[width=0.210\textwidth]{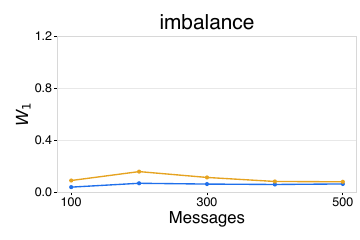} \\
\includegraphics[width=0.210\textwidth]{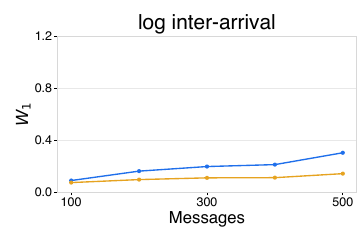} &
\includegraphics[width=0.210\textwidth]{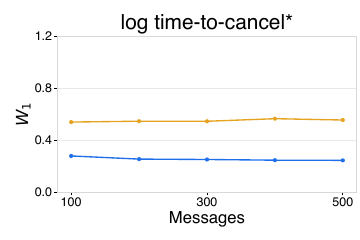} &
\includegraphics[width=0.210\textwidth]{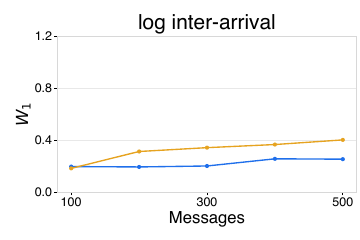} &
\includegraphics[width=0.210\textwidth]{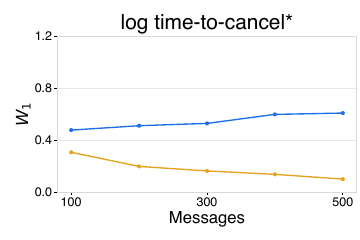} \\
\includegraphics[width=0.210\textwidth]{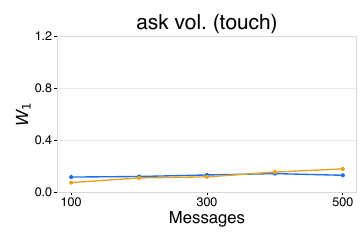} &
\includegraphics[width=0.210\textwidth]{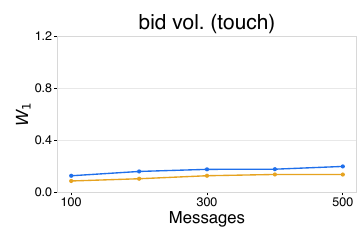} &
\includegraphics[width=0.210\textwidth]{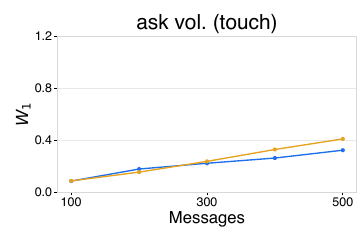} &
\includegraphics[width=0.210\textwidth]{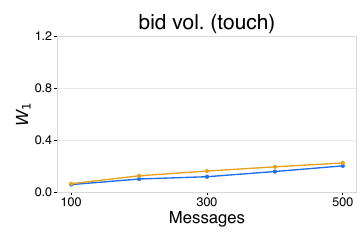} \\
\includegraphics[width=0.210\textwidth]{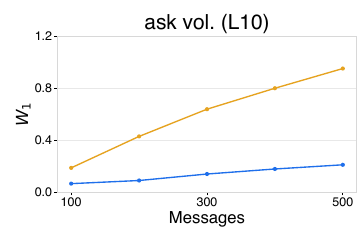} &
\includegraphics[width=0.210\textwidth]{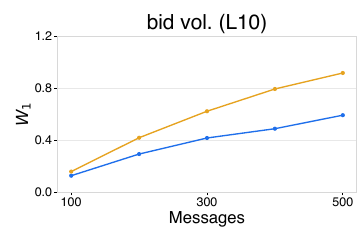} &
\includegraphics[width=0.210\textwidth]{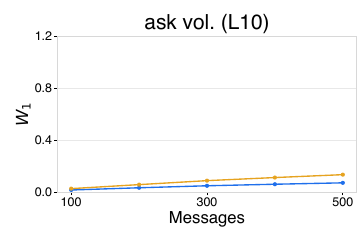} &
\includegraphics[width=0.210\textwidth]{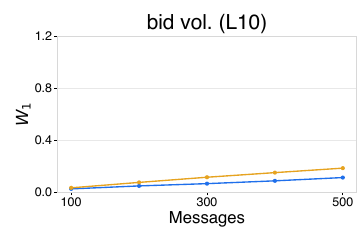} \\
\includegraphics[width=0.210\textwidth]{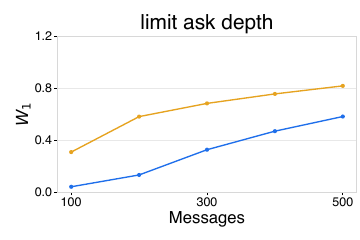} &
\includegraphics[width=0.210\textwidth]{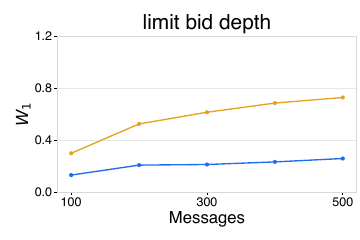} &
\includegraphics[width=0.210\textwidth]{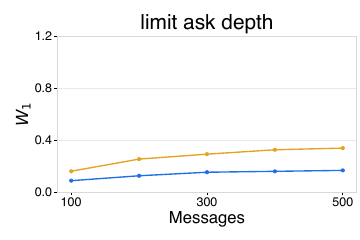} &
\includegraphics[width=0.210\textwidth]{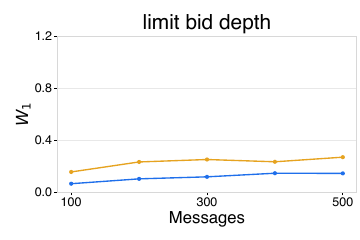} \\
\includegraphics[width=0.210\textwidth]{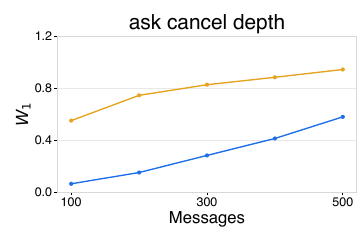} &
\includegraphics[width=0.210\textwidth]{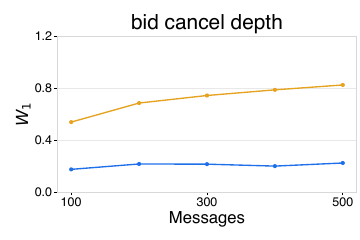} &
\includegraphics[width=0.210\textwidth]{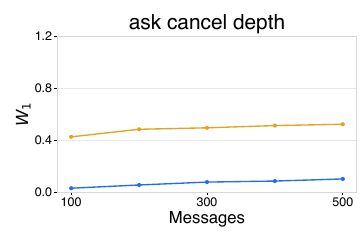} &
\includegraphics[width=0.210\textwidth]{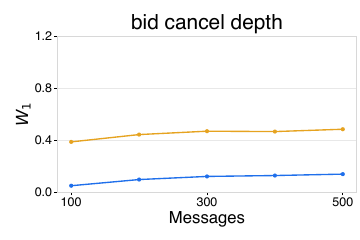} \\
\includegraphics[width=0.210\textwidth]{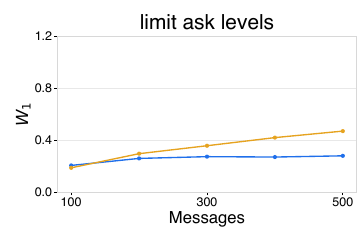} &
\includegraphics[width=0.210\textwidth]{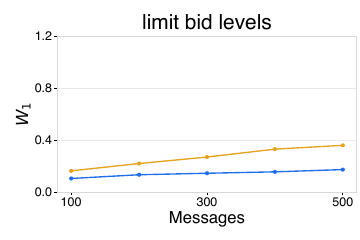} &
\includegraphics[width=0.210\textwidth]{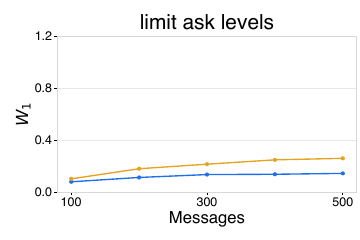} &
\includegraphics[width=0.210\textwidth]{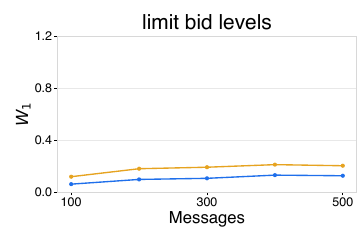} \\
\includegraphics[width=0.210\textwidth]{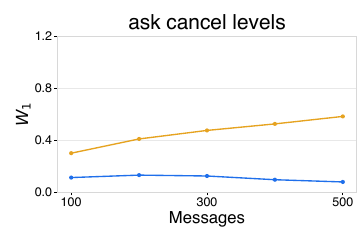} &
\includegraphics[width=0.210\textwidth]{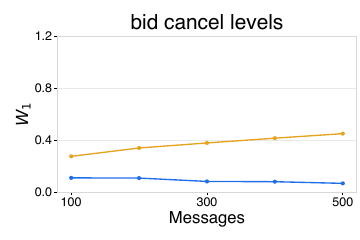} &
\includegraphics[width=0.210\textwidth]{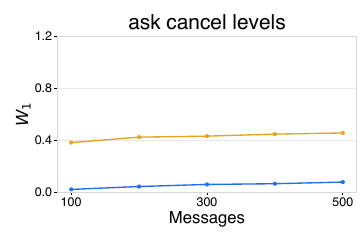} &
\includegraphics[width=0.210\textwidth]{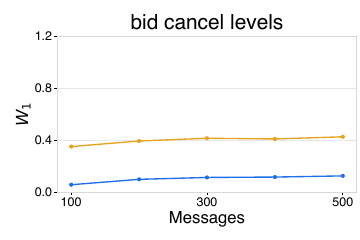} \\
\includegraphics[width=0.210\textwidth]{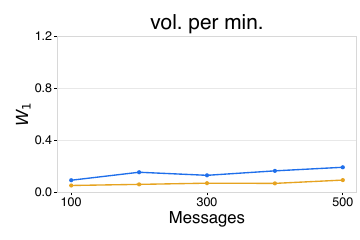} &
& \includegraphics[width=0.210\textwidth]{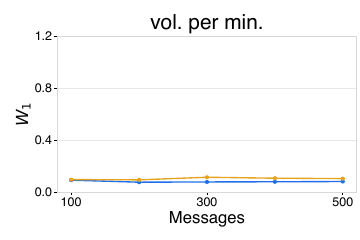} &
\end{tabular}
\caption{\textbf{LOB-Bench evaluation: error accumulation in Wasserstein distances.}
Log time-to-cancel* measures time-to-cancel for all orders canceled during the rollout using exact resting-order information, including cancellations of orders present in the initial LOB that were excluded from the original LOB-Bench implementation.
Lower is better.}
\label{fig:lob_bench_divergence_w1}
\end{figure}

\clearpage
\paragraph{Error accumulation.}
\cref{fig:lob_bench_divergence_l1,fig:lob_bench_divergence_w1} show $L_1$ and Wasserstein divergences between generated and real distributions on GOOG and INTC at 100-message intervals.
ReLOBGen generally exhibits lower divergence values and flatter slopes than LOBS5 over the rollout horizon.

\clearpage
\subsection{Marginal event-type distributions}
\label{app:marginal_event_distribution}

\cref{tab:marginal_event_distribution} compares marginal event-type distributions in actual and generated messages on GOOG and INTC.
ReLOBGen achieves the lowest total variation distance among the evaluated methods on both stocks, indicating a closer match to the actual event-type distribution.


\begin{table}[H]
  \centering
  \caption{\textbf{Marginal event-type frequencies in actual and generated messages.}
  TV denotes the total variation distance from the actual distribution, reported in percentage points.
  Lower TV is better, and bold indicates the lowest value.}
  \label{tab:marginal_event_distribution}
  \renewcommand{\arraystretch}{1.1}
  \setlength{\tabcolsep}{5pt}
  \scalebox{0.80}{
    \begin{tabular}{@{}llccccc}
      \toprule
      Ticker & Method & Add & Cancel & Delete & Execute & TV \\
      \midrule
      \multirow{4}{*}{GOOG} & Actual & 50.2\% & 0.2\% & 46.6\% & 3.0\% & -- \\ \cmidrule(lr){2-7}
      & LOBS5~(ref.-last)  & 51.0\% & 0.3\% & 36.9\% & 11.8\% & 9.6 pp \\
      & Ref.-first + uniform & 45.2\% & 0.2\% & 38.4\% & 16.2\% & 13.2 pp \\
      & \cellcolor{gray!10}ReLOBGen & \cellcolor{gray!10}48.7\% & \cellcolor{gray!10}0.2\% & \cellcolor{gray!10}41.9\% & \cellcolor{gray!10}9.2\% & \cellcolor{gray!10}\textbf{6.2 pp} \\
      \midrule
      \multirow{4}{*}{INTC} & Actual & 49.7\% & 0.2\% & 43.7\% & 6.4\% & -- \\ \cmidrule(lr){2-7}
      & LOBS5~(ref.-last)  & 52.5\% & 0.1\% & 43.4\% & 4.0\% & 2.8 pp \\
      & Ref.-first + uniform & 47.4\% & 0.2\% & 44.1\% & 8.3\% & 2.3 pp \\
      & \cellcolor{gray!10}ReLOBGen & \cellcolor{gray!10}50.8\% & \cellcolor{gray!10}0.2\% & \cellcolor{gray!10}44.8\% & \cellcolor{gray!10}4.2\% & \cellcolor{gray!10}\textbf{2.2 pp} \\
      \bottomrule
    \end{tabular}
  }
\end{table}




\clearpage
\section{Resting-order coverage by history length}
\label{app:resting_order_coverage}

\begin{figure}[H]
    \centering
    \includegraphics[width=\textwidth]{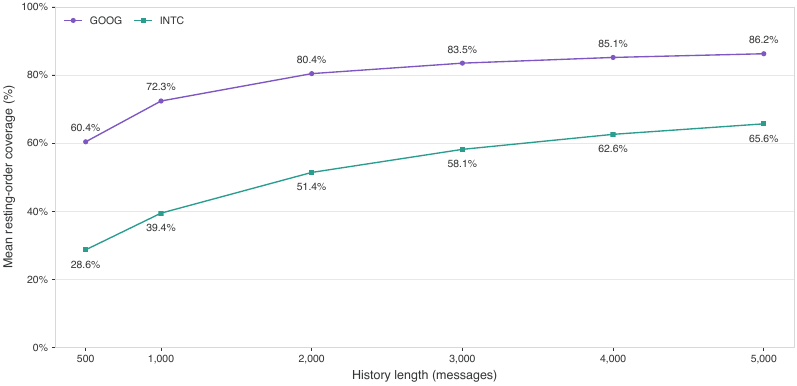}
    \caption{\textbf{Mean resting-order coverage by history length on GOOG and INTC.}
    Coverage is evaluated during regular trading hours on the test split.}
    \label{fig:resting_order_coverage}
\end{figure}

\cref{fig:resting_order_coverage} shows resting-order coverage as a function of model-history length for GOOG and INTC.
Let $\mathcal O_{t-1}^{(10)}$ denote the subset of $\mathcal O_{t-1}$ within the current top 10 price levels on each side.
We define coverage as
\begin{align}
    C_t(L) = \frac{\left|\left\{o \in \mathcal O_{t-1}^{(10)} : o \text{ appears in } M_{t-L:t-1}\right\}\right|}{|\mathcal O_{t-1}^{(10)}|},
\end{align}
and report the mean of $C_t(L)$ across windows.
An order appears in the history if its raw order ID occurs in the preceding $L$ filtered messages.

At the 500-message context length used by our models, coverage is only 60.4\% for GOOG and 28.6\% for INTC.
Coverage increases to 86.2\% and 65.6\%, respectively, with 5,000 preceding messages, with diminishing gains as history length grows.
Even at this length, a substantial fraction of the current resting orders remains absent from the message history.

\end{document}